\documentclass[british,10pt,twocolumn,letterpaper]{article}
\usepackage[T1]{fontenc}
\usepackage[latin9]{luainputenc}
\usepackage{array}
\usepackage{units}
\usepackage{multirow}
\usepackage{amsmath}
\usepackage{amssymb}
\usepackage{graphicx}
\usepackage{amsthm}

\makeatletter

\providecommand{\tabularnewline}{\\}

\newtheorem{theorem}{Theorem}

\newtheorem{corollary}{Corollary}

\usepackage{color,soul}
\definecolor{orange}{rgb}{1,0.6,0.3}
\definecolor{dblue}{rgb}{0.80,0.85,0.9}
\setulcolor{orange}
\setul{0em}{0.15em}
\sethlcolor{dblue}

\definecolor{Gray}{gray}{0.9}

\usepackage{float}

\usepackage{iccv}
\usepackage{amssymb}

\usepackage{times}
\usepackage{epsfig}
\usepackage{graphicx}
\usepackage{amsmath}

\usepackage{placeins}
\usepackage{color}
\usepackage{colortbl}
\usepackage{rotating}
\definecolor{lightgray}{gray}{0.8}
\definecolor{verylightgray}{gray}{0.9}

\usepackage[pagebackref=false,breaklinks=true,letterpaper=true,colorlinks,bookmarks=false]{hyperref}

\def\iccvPaperID{0489} 

\iccvfinalcopy
\ificcvfinal\fi

\@ifundefined{showcaptionsetup}{}{%
 \PassOptionsToPackage{caption=false}{subfig}}
\usepackage{subfig}
\makeatother

\usepackage{babel}
\begin{document}
\makeatletter  
\renewcommand{\paragraph}{%
\@startsection{paragraph}{4}%
 {\z@}{0.5ex \@plus 1ex \@minus .2ex}{-0.5em}%
  {\normalfont \normalsize \bfseries}%
} 

\let\originalparagraph\paragraph 
\renewcommand{\paragraph}[2][.]{\originalparagraph{#2#1}}

\makeatother

\vspace{0em}

\title{Adversarial Image Perturbation for Privacy Protection \\ A Game Theory Perspective}

\author{Seong Joon Oh\qquad{}Mario Fritz\qquad{}Bernt Schiele\vspace{0em}
\\
\begin{tabular}{c}
\normalsize{Max Planck Institute for Informatics}, 
\normalsize{Saarland Informatics Campus}, 
\normalsize{Saarbr\"ucken, Germany}\tabularnewline
\texttt{\small{}\{joon,mfritz,schiele\}@mpi-inf.mpg.de}\tabularnewline
\end{tabular}
}
\maketitle

\begin{abstract}
Users like sharing personal photos with others through social media. At the same time, they might want to make automatic identification in such photos difficult or even impossible. Classic obfuscation methods such as blurring are not only unpleasant but also not as effective as one would expect \cite{joon16eccv,Wilber2016avoid,mcpherson2016defeating}. Recent studies on \emph{adversarial image perturbations} (AIP) suggest that it is possible to confuse recognition systems effectively without unpleasant artifacts. However, in the presence of counter measures against AIPs \cite{graese2016icmla}, it is unclear how effective AIP would be in particular when the choice of counter measure is unknown. Game theory provides tools for studying the interaction between agents with uncertainties in the strategies. We introduce a general game theoretical framework for the user-recogniser dynamics, and present a case study that involves current state of the art AIP and person recognition techniques. We derive the optimal strategy for the user that assures an upper bound on the recognition rate independent of the recogniser's counter measure. Code is available at \url{https://goo.gl/hgvbNK}.
\end{abstract}

\section{\label{sec:Introduction}Introduction}

\noindent
People nowadays share massive amounts of personal photos through social media. Personal photos contain rich private information, \eg about family members, travel destinations, and political activities. Together with recent developments in computer vision techniques \cite{Deng2009CvprImageNet,Krizhevsky2012Nips,He2016DeepRL,Oh2015Iccv,sun2017cvpr}, this results in increasing concerns that malicious entities employing computer vision technologies could extract private information from visual data. 

Classical obfuscation techniques, such as face blurring and pixellisation, is not only unpleasant but also ineffective against convnet-based recognisers \cite{joon16eccv,Wilber2016avoid,mcpherson2016defeating}. 

There have been recent studies on \emph{adversarial image perturbations} (AIP): carefully crafted additive perturbations on the image that confuses a convnet while being nearly invisible to human eyes \cite{szegedy2014iclr,goodfellow2015iclr,moosavi2016cvpr,moosavi2016universal}. AIPs are indeed promising as obfuscation techniques.

\begin{figure}
\begin{centering}
\includegraphics[width=1\columnwidth]{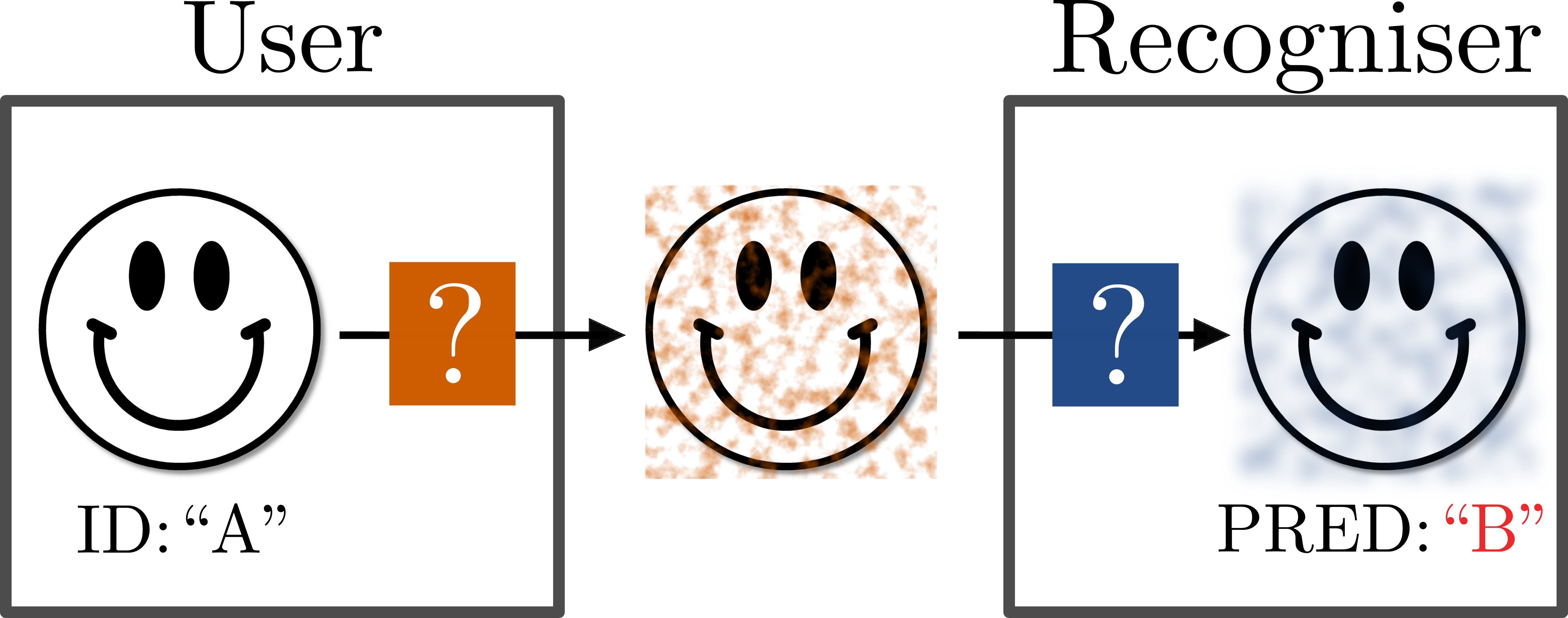}
\par\end{centering}
\caption{\label{fig:teaser}A game between a social media user and a recogniser over a photo. The user perturbs the image using orange strategy, trying to confuse the recogniser. The recogniser chooses blue strategy as a counter measure. They do not know which strategy is picked by the other.}
\end{figure}

However, it remains a question whether AIPs are still effective when counter measures are taken. For example, \cite{graese2016icmla} proposed simple image processing tactics to counter the AIP effects (\eg blurring by small amount). If furthermore the particular choice of counter measure is unknown, the best strategy is not obvious for the user.

Game theory provides useful tools for analysis when there exist uncertainties in the strategies for each player. We present a game theoretical framework to describe a system in which the user and recogniser strive for antagonistic goals: dis-/enabling recognition. This framework makes it possible to derive guarantees on the user's level of privacy, independent of the recogniser's counter measure, from an explicitly formulated set of assumptions. We include a case study of a person identification game, deriving the user's privacy guarantee with respect to the current state of the art AIP and person recognition methods.

This paper showcases the utility of game theory in understanding the user-recogniser dynamics. The framework can be extended beyond the particular settings considered. We believe this framework will further aid user-recogniser analyses in more diverse tasks and setups.

We list our contributions as follows:

\begin{itemize}
\item A game theoretic framework for studying the user-recogniser dynamics.
\item Application of \emph{adversarial image perturbation} (AIP) as an effective and aesthetic technique for person obfuscation.
\item Novel robust and recogniser-selective AIPs.
\item An empirical case study of the game theoretic framework, leading to the privacy guarantees for the user.
\end{itemize}


\section{\label{sec:Related-work}Related Work}

\paragraph{Privacy and computer vision}

While there exists a bulk of research on user privacy traditionally led by the security community \cite{Narayanan2009SspDeAnonymizingFacebook,Zheleva2009Icwww,Narayanan2010Cacm,Mislove2010}, studies on private content in visual data began only recently \cite{Wilber2016avoid,joon16eccv,mcpherson2016defeating}.

Wilber \etal \cite{Wilber2016avoid} studied the performance of a commercial \emph{face detector} under multiple face obfuscation methods (blur, darkening, camouflage glasses, etc.). Oh \etal \cite{joon16eccv} and McPherson \etal \cite{mcpherson2016defeating} studied the \emph{face recognition} performance. In particular, \cite{joon16eccv} showed that current recognisers can adapt to obfuscation patterns. Above works conclude that recognisers can be robust against simple obfuscation methods like face blurring. In this work, we study a stronger obfuscation type: adversarial image perturbations.

\paragraph{Adversarial image perturbation (AIP)}

Szegedy \etal \cite{szegedy2014iclr} first studied the phenomenon of adversarial instability of convnets: it is possible to generate invisible additive perturbations that completely fool a recogniser. The initial crafting algorithm was based on the L-BFGS \cite{szegedy2014iclr}; more efficient first-order algorithms have been proposed \cite{goodfellow2015iclr,Rozsa2016AdversarialDA,moosavi2016cvpr,kurakin2016adversarial}. We review existing AIP algorithms and our novel variants conceptually and empirically.

\paragraph{Robust classification against AIPs}

Some pre-convnet works considered enhancing general robustness of classifiers by training on adversarial data. Lanckriet \etal \cite{Lanckriet2003minimax} trained a linear classifier on adversarial data constrained to a fixed mean and covariance for each class. Br\"{u}ckner \etal \cite{Bruckner2012NashML} introduced game theoretic concepts to formalise the adversarial training procedure. However, they limited their attention to simpler models: linear \cite{Lanckriet2003minimax} or convex \cite{Bruckner2012NashML}. This work builds on a game theoretic framework which accommodates state of the art convnet models.

Since the advent of effective convnets \cite{Krizhevsky2012Nips} and corresponding AIP algorithms \cite{szegedy2014iclr}, some works \cite{goodfellow2015iclr,Huang2015LearningWA} have considered training convnets with AIPs, achieving robustness against AIPs to some extent. On the other hand, Graese \etal \cite{graese2016icmla} argued that simple test time image processing, such as translation, Gaussian noise, blurring, and re-sizing, can equally neutralise the effect of AIPs, without having to re-train the convnet. In our case study, we include those image processing methods in the recogniser's strategy space.

\paragraph{Robust AIPs against classifiers}

Sharif \etal \cite{Sharif16AdvML} proposed a method for robustification by optimising an AIP against a set of images, rather than a single image. This approach was also suggested by Moosavi \etal \cite{moosavi2016universal} for generating \emph{universal perturbations}. In our work, we consider optimising the AIP against a set of jittered versions of the target input. We will show empirically that this enables a targetted defense against image processing strategies.

\paragraph{AIP for identity obfuscation}

This paper advocates the AIP as an effective and aesthetic means for disabling recognition. Previously Sharif \etal \cite{Sharif16AdvML} also used adversarial optimisation to fool a person recogniser. Compared to their limited setup (fixed pose, fixed recognition strategy), our case study covers a large-scale social media setup with user-recogniser dynamics.

\paragraph{Person recognition task}

Our case study considers the person recognition task in social media setup \cite{Gallagher2008Cvpr,Zhang2015CvprPiper,Oh2015Iccv}, as opposed to face recognition \cite{Huang2007Lfw} (frontal faces, good lighting) or pedestrian re-identification \cite{Benfold2009BmvcTownCentre,Bedagkar2014IvcPersonReIdSurvey} (low resolution, fixed context). Social media photos capture subjects appearing in diverse range of viewpoints, poses, clothings, and events. Zhang \etal introduced PIPA \cite{Zhang2015CvprPiper}, the first large-scale social media person recognition dataset and benchmark. Our empirical studies are built upon this dataset.

\paragraph{Person recognition models}

Multiple researchers have proposed person recognition techniques
in social media photos. Zhang \etal \cite{Zhang2015CvprPiper} proposed
to combine cues from multiple body parts obtained by poselet detections.
Oh \etal \cite{Oh2015Iccv} greatly simplified \cite{Zhang2015CvprPiper}  while achieving the state of the art performance. We build our recogniser model upon \cite{Oh2015Iccv}, possibly with more advanced network architectures. A concurrent work by Liu \etal \cite{liu2017learning} claims to have improved the method via metric learning objective. There exist other works \cite{Li_2016_CVPR,joon16eccv,Li_2017_CVPR}, which exploit social media metadata.

\begin{figure*}
\begin{centering}
\includegraphics[width=1.8\columnwidth]{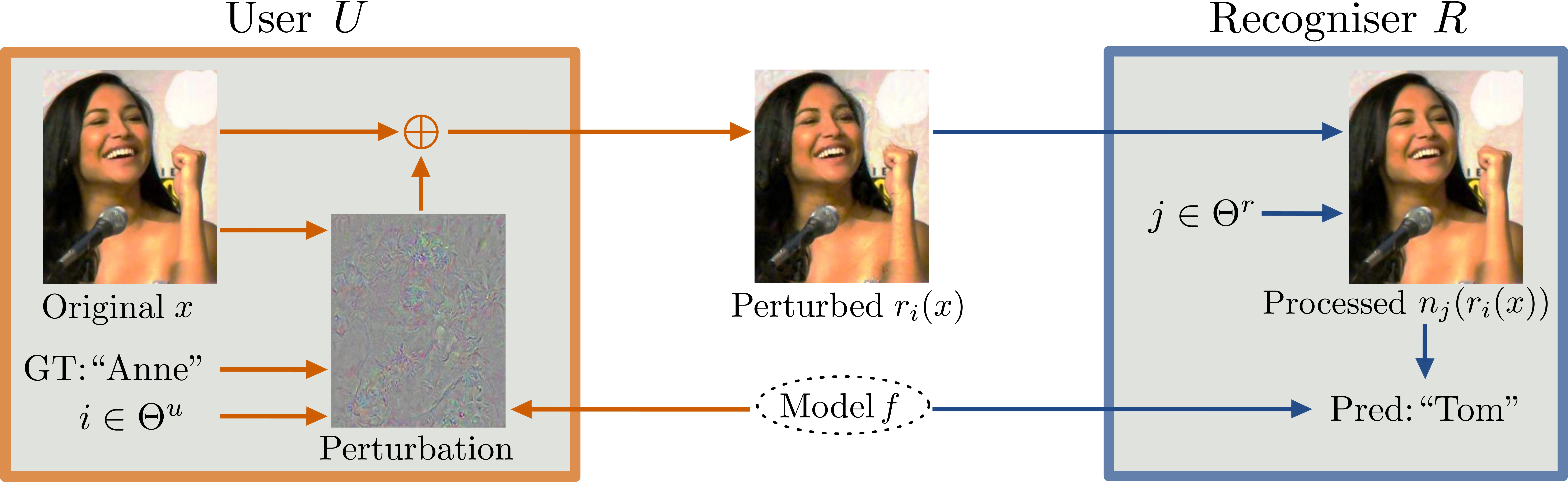}
\par\end{centering}
\caption{\label{fig:setup}User-recogniser game on a single photo. Each player does not know the opponent's strategy. Orange (blue) arrows indicate actions taken by the user (recogniser).  Information in the orange (blue) box is only available to the user (recogniser).}
\end{figure*}

\section{\label{sec:Scenarios}User-Recogniser Game}

\noindent
This section provides a general framework for studying user-recogniser games. The framework provides a tool for systemising the path from a set of explicit assumptions on the players to game theoretical conclusions.

Our user-recogniser game framework is visualised in figure \ref{fig:setup}. The user $U$ perturbs the original image $x$ according to a strategy $i\in\Theta^u$, aiming to thwart recognition. The recogniser $R$ processes the perturbed image $r_i(x)$ according to a strategy $j \in \Theta^r$, aiming to neutralise the effect of image perturbation. The resulting image $n_j(r_i(x))$ is passed to the model $f$ to make a prediction. The game arises from the fact that each player does not know the opponent's strategy, although they do know each other's strategy space. 

We introduce relevant game theoretical concepts and key theoretical results in \S \ref{sec:Game-Theory} to help formalise the framework in \S \ref{sec:Framework}. We discuss possible extensions in \S \ref{sec:limitations}.

\subsection{\label{sec:Game-Theory}Two-Person Constant-Sum Games}

\noindent
We describe our system as a {\bf two-person game} \cite{Neumann1928} consisting of two players, the user $U$ and the recogniser $R$ with designated {\bf strategy spaces}, $\Theta^u$ and $\Theta^r$.

As a result of each player committing to strategies $i \in \Theta^u$ and $j \in \Theta^r$ respectively, $R$ receives a {\bf payoff} of $p_{ij}$, the recognition rate; $U$ then receives a payoff of $1-p_{ij}$, the mis-recognition rate.

Game theory suggests that it is sometimes better to randomise the strategies. $U$ can adopt a {\bf mixed (random) strategy} $\theta^u=(\theta^u_i)_{i\in\Theta^u}$, defined as a distribution over the strategy space $\Theta^u$, and similarly for $R$. With abuse of notation we write $p(\theta^u,\theta^r):=\underset{i,j}{\sum}\theta^u_i \theta^r_j p_{ij}$ for the expected payoff for $R$ when the mixed strategies $\theta^u$ and $\theta^r$ are taken. The payoff for $U$ is derived and defined as $\underset{i,j}{\sum}\theta^u_i \theta^r_j (1-p_{ij})=1-p(\theta^u,\theta^r)=:p^\prime(\theta^u,\theta^r)$.

We say that a two-person game is a {\bf constant-sum game} if the players' payoffs sum to a constant $\beta$ independent of the strategies. In our case, the recognition and mis-recognition rates always sum to one ($\beta=1$). A game is {\bf finite} if the strategy spaces are finite. We have the following optimality theorem.

\begin{theorem}[von Neumann \cite{Neumann1928}, 1928]
\label{thm:von-neumann}
For a finite constant-sum game, there exist {\bf \emph{optimal}} or {\bf \emph{minimax}} mixed strategies $\theta^{u\star}$ and $\theta^{r\star}$ such that 
\begin{equation}
\label{eq:optimal-strategy}
p(\theta^{u\star},\theta^{r})\leq p(\theta^{u\star},\theta^{r\star})\leq p(\theta^{u},\theta^{r\star}) \,\,\,\,\,\, \forall \, \theta^u, \theta^r
\end{equation}
where $v:=p(\theta^{u\star},\theta^{r\star})$ is the {\bf \emph{value of the game}}.
\end{theorem}

\noindent
Equation \ref{eq:optimal-strategy} implies that when $R$ plays $\theta^{r\star}$, $R$ is guaranteed to have a payoff of at least $v$, regardless of $U$'s strategy; if $U$ plays $\theta^{u\star}$, $U$ is guaranteed to have a payoff of $1-v$. In our scenario, this means that $U$'s optimal strategy guarantees a certain mis-recognition rate, regardless of $R$'s strategy. 

$U$'s optimal strategies can be obtained efficiently via linear programming that solves the following ($R$'s optimal strategy can be found by swapping min and max):
\begin{equation}
\label{eq:minimax-optimisation}
\underset{\theta^{u}}{\arg\min}\,\, \underset{\theta^{r}}{\max}\,\, \underset{i,j}{\sum}\theta^u_i \theta^r_j p_{ij} \,\,\,\, \text{s.t.} \,\,\, \theta^u,\,\theta^r\text{ are distributions.}
\end{equation}
If $U$ has knowledge on $R$'s strategy $\bar{\theta^r}$, then $U$ can take advantage of this knowledge. $U$ can optimise her strategy given $\bar{\theta^r}$ to attain a payoff of $\underset{\theta^u}{\max}\,\, p^\prime(\theta^u,\bar{\theta^r})\geq p^\prime(\theta^{u\star},\bar{\theta^r})\geq p^\prime(\theta^{u\star},\theta^{r\star})=1-v$, a potentially better payoff than the no-knowledge scenario $1-v$. However, if $R$'s strategy is optimal $\bar{\theta^r}=\theta^{r\star}$, then the knowledge does not bring improvement for $U$: $\underset{\theta^u}{\max}\,\, p^\prime(\theta^u,\theta^{r\star})=1-v$.

In reality, not all players play optimally either due to the lack of knowledge (\eg on the opponent's strategy space), or due to pure irrationality. We refer to such a player as an {\bf irrational player}. Our discussion above implies:
\begin{corollary}[] If $U$ knows $R$'s strategy $\bar{\theta^r}$, and if it is suboptimal, then $U$ can enjoy a better payoff than $1-v$.
\end{corollary}

\subsection{\label{sec:Framework}Components of the User-Recogniser Game}

\noindent
 We specify the payoffs, strategy spaces, and information allowed for the user $U$ and the recogniser $R$.

\paragraph{Test data}

We assume that the test data are distributed according to $(\hat{x},\hat{y})\sim D$. This dataset is the source of private information that the two players compete for.

\paragraph{Fixed model}

We assume that $U$ and $R$ use a fixed model $f$ (\eg a publicly available model).
This is a reasonable assumption, as $U$ and $R$ often would not have resources to train modern convnets.

\paragraph{Known model}

Each player is aware that the opponent uses $f$. This may be unrealistic, but provides a good starting point. Relaxation of this assumption is discussed in \S \ref{sec:limitations}.

\paragraph{Payoff}

When the players commit to strategies $i\in\Theta^u$ and $j\in\Theta^r$, $R$'s payoff is the recognition rate on the test set:
\vspace{-0.5em}
\begin{equation}
\label{eq:payoff}
p_{ij}=
\underset{(\hat{x},\hat{y})\sim D}{\mathbb{P}}\left[
\underset{y}{\arg\max}\,\, f^{y}\left(n_j\left(r_i\left(\hat{x}\right)\right)\right)=\hat{y}
\right]
\vspace{-0.5em}
\end{equation}
where $f^y$ denotes the model prediction score for class $y$. $U$ receives the payoff $1-p_{ij}$, the mis-recognition rate.

\paragraph{User's strategy space {$\Theta^u$}}

We consider additive perturbations such that for an input $x$,
\vspace{-0.5em}
\begin{equation}
\label{eq:L2-constraint}
r_{i}\left(x\right)=x+t(x),\qquad ||t(x)||_2\leq\epsilon
\vspace{-0.5em}
\end{equation}
for some constant $\epsilon>0$. When $\epsilon$ is small enough, the perturbation is nearly invisible to human eyes (see figure \ref{fig:example-adversarial-perturbation}). These perturbations are frequently referred to as \emph{adversarial image perturbations} (AIPs). We discuss existing AIPs and our novel variants in \S \ref{sec:Adversarial}.

\paragraph{Recogniser's strategy space {$\Theta^r$}}

$R$ aims to neutralise the adversarial effect of AIPs. Although some works have suggested re-training the model with AIPs, demonstrating certain degree of robustification \cite{goodfellow2015iclr,Huang2015LearningWA}, Graese \etal \cite{graese2016icmla} has argued that simple image processing can already neutralise the AIP effects cheaply and effectively. They have demonstrated that on MNIST, translation ($\mathsf{T}$), Gaussian additive noise ($\mathsf{N}$), blurring ($\mathsf{B}$), and cropping \& re-sizing ($\mathsf{C}$) have improved the recognition rate from 0\% (post-AIP) to 68\%, 58\%, 65\%, and 76\%, respectively. In our case study, we will include these transformations in $\Theta^r$. In \S \ref{sec:limitations}, we will discuss about expanding strategy spaces.

\paragraph{Known strategy spaces}

The strategy spaces for each player ($\Theta^u$ and $\Theta^r$) are known to each other, while the chosen strategies are not known.

\paragraph{Multiple recognisers}

$U$ may encounter a set of recognisers not all of which are malicious. For example, $U$ uploads her personal photos to a cloud service with a recognition system $R_1$; she wants an AIP that enables a successful recognition by $R_1$ but disables recognition by a malicious system $R_2$. We propose an approach for generating \emph{selective} AIPs in \S \ref{sec:our-aips} and confirm their existence in \S \ref{sec:Multi-Agent}. From a theoretical standpoint, the existence of selective AIPs attest to the diversity of possible AIP patterns, in line with the existence of \emph{universal perturbations} \cite{moosavi2016universal}.

\subsection{\label{sec:limitations}Extensions}

\noindent
In the previous section, we have introduced the user-recogniser game framework with particular assumptions explored in this paper. In this section, we show that the framework can be extended beyond this setup.

\paragraph{Unknown models}

Many AIP techniques assume a full knowledge on the model $f$, but the computation of \emph{black-box} AIPs is another active research field \cite{papernot2016transferability,papernot2017practical,narodytska2016simple,Liu2016DelvingIT}; $U$ can potentially adopt these methods.

\paragraph{Non-constant sum}

If $U$ and $R$ assign different weights to different test samples, then the payoffs may not sum to 1. For such non-constant sum games, there exist \emph{Nash equilibrium} strategies for each player \cite{Nash50}. The optimal strategy and payoff analyses are still possible.

\paragraph{Non-additive AIPs}

The framework allows $r_i$ to be any function that induces invisible changes on the image. Current restriction to equation \ref{eq:L2-constraint} rules out \eg one-pixel translation of the whole image. Most, if not all, prior work on AIP is done in the additive setup. Crafting non-additive AIP would be interesting future work.

\paragraph{Non-fixed models}

$R$ with enough computational resources may re-train the model $f$ with AIPs. One option to expand our framework to such a setup would be to incorporate the model parameters in $\Theta^r$. Br\"{u}ckner \etal \cite{Bruckner2012NashML} have studied this setup, but have assumed convex loss functions. Understanding games with continuous strategy spaces and non-convex payoffs (\eg convnet losses) is an open question both for computer vision and game theory research.

\paragraph{Unknown strategy spaces}

The exact possible set of strategies may not be known to the opponent. With improving technologies, the respective strategy spaces may even grow over time. The framework cannot do much about the unknown strategies, but can adaptively expand the strategy spaces according to technological developments.

\section{\label{sec:Adversarial}Adversarial Image Perturbation Strategies}

\noindent
This section reviews existing adversarial image perturbation (AIP) algorithms that use first-order optimisation schemes, and proposes our novel variants.

We compute AIPs as additive transformations with $L_2$ norm constraints (equation \ref{eq:L2-constraint}). Computation of AIP can be formulated as a loss \emph{maximisation} problem
\vspace{-0.5em}
\begin{equation}
\label{eq:AIP-objective}
\underset{t}{\max}\,\, \mathcal{L}\left(f\left(x+t\right),y\right)
\qquad \text{s.t.} \,\,\, ||t||_2\leq\epsilon
\vspace{-0.5em}
\end{equation}
\noindent
where $x$ is the input image and $y$ is the ground truth label; the loss function $\mathcal{L}$ is to be specified.

\subsection{\label{sec:exising-aips}Existing AIP methods}

\noindent
Depending on the loss function $\mathcal{L}$ and the optimisation algorithm, we recover most of the existing AIP methods such as Fast Gradient Vector \cite{Rozsa2016AdversarialDA}, Fast Gradient Sign \cite{goodfellow2015iclr}, Basic Iterative \cite{kurakin2016adversarial}, and DeepFool \cite{moosavi2016cvpr}. The \emph{universal perturbations} introduced by Moosavi \etal \cite{moosavi2016universal} can also be seen as a special case of equation \ref{eq:AIP-objective} where the loss is computed over the entire test set and the perturbation $t$ is shared across images. See table \ref{tab:review-adversarial-perturbation} for the summary.

\begin{table}
\begin{centering}
\begin{tabular}{ccccc}
\multirow{2}{*}{Variants} &  & \multirow{2}{*}{Loss $\mathcal{L}$} & Stopping  & \multirow{2}{*}{Step size}\tabularnewline
 &  &  & condition & \tabularnewline
\vspace{-1em}
 &  &  &  & \tabularnewline
\cline{1-1} \cline{3-5} 
\vspace{-1em}
 &  &  &  & \tabularnewline
\cline{1-1} \cline{3-5} 
\vspace{-1em}
 &  &  &  & \tabularnewline
$\mathtt{FGS}$\cite{goodfellow2015iclr} &  & $-\log \hat{f}^{y}$ & 1 iteration & Fixed\tabularnewline
\vspace{-1em}
 &  &  &  & \tabularnewline
$\mathtt{FGV}$\cite{Rozsa2016AdversarialDA} &  & $-\log \hat{f}^{y}$ & 1 iteration & Fixed\tabularnewline
\vspace{-1em}
 &  &  &  & \tabularnewline
\cline{1-1} \cline{3-5} 
\vspace{-1em}
 &  &  &  & \tabularnewline
$\mathtt{BI}$\cite{kurakin2016adversarial} &  & $-\log \hat{f}^y$ & $K$ iterations & Fixed\tabularnewline
\vspace{-1em}
 &  &  &  & \tabularnewline
$\mathtt{GA}$ &  & $-\log \hat{f}^y$ & $K$ iterations & Fixed\tabularnewline
\vspace{-1em}
 &  &  &  & \tabularnewline
\cline{1-1} \cline{3-5} 
\vspace{-1em}
 &  &  &  & \tabularnewline
 $\mathtt{DF}$\cite{moosavi2016cvpr} &  & $f^{y^{c}}-f^{{y}}$ & $K$ it.$\vee$ fooled & Adaptive\tabularnewline
\vspace{-1em}
 &  &  &  & \tabularnewline
$\mathtt{GAMAN}$ &  & $f^{y^{\star}}-f^{{y}}$ & $K$ iterations & Fixed\tabularnewline
\vspace{-1em}
 &  &  &  & \tabularnewline
\cline{1-1} \cline{3-5} 
\end{tabular}
\par\end{centering}
\vspace{0em}

\caption{\label{tab:review-adversarial-perturbation}Conceptual differences
among AIP methods. $f^{y^\prime}$ is the model score for class $y^\prime$, and $\hat{f}$ denotes the softmax output of $f$. $y$ is the ground truth label, and $y^{\star}$ is the most likely label among wrong ones. $y^{c}$ is the label with the closest linearised decision boundary.}
\end{table}

\paragraph{Fast Gradient Vector ($\mathtt{FGV}$) \cite{Rozsa2016AdversarialDA}}

$\mathtt{FGV}$ adopts the softmax-log loss $\mathcal{L}=-\log \hat{f}^y$ in equation \ref{eq:AIP-objective}, solving it via one-step gradient ascent: $t^\star=-\gamma\nabla \mathcal{L}(x)$ for some constant $\gamma>0$.

\paragraph{Fast Gradient Sign ($\mathtt{FGS}$) \cite{goodfellow2015iclr}}

$\mathtt{FGS}$ is identical to $\mathtt{FGV}$, except that $\nabla \mathcal{L}(x)$ is replaced with $\text{sign}\left(\nabla \mathcal{L}(x)\right)$.

\paragraph{Gradient Ascent ($\mathtt{GA}$)}

This is a multi-step variant of $\mathtt{FGV}$. Perturbation is initialised at $t^{(0)}=0$. Gradient ascent is performed on the loss function iteratively: $t^{(m+1)}=t^{(m)}-\gamma\nabla\mathcal{L}(x+t^{(m)})$ for $m=0,\cdots,K$ for some fixed step size $\gamma>0$ and maximal number of iterations $K\geq 1$.

\paragraph{Basic Iterative ($\mathtt{BI}$) \cite{kurakin2016adversarial}}

$\mathtt{BI}$ is identical to $\mathtt{GA}$, except that $\nabla \mathcal{L}(x)$ is replaced with $\text{sign}\left(\nabla \mathcal{L}(x)\right)$.

\paragraph{DeepFool ($\mathtt{DF}$) \cite{moosavi2016cvpr}}

$\mathtt{DF}$ algorithm solves the objective: 
\vspace{-0.5em}
\begin{equation}
\label{eq:deepfool-objective}
\underset{t}{\min}\,\,||t||_{2}\quad\text{s.t.}\quad\underset{y}{\arg\max}\,\,f^{y}\left(x+t\right)\neq y
\vspace{-0.5em}
\end{equation}
which finds the minimal perturbation such that the prediction is wrong. Although the objective is different, we show that the $\mathtt{DF}$ algorithm can also be seen as a first-order method solving equation \ref{eq:AIP-objective} for some loss function.

$\mathtt{DF}$ first finds the class with the nearest decision hyperplane, denoted by $c$. To simplify the search, $c$ is found on the linear approximation of $f$ around $x$ (tangent function). The normal vector to the decision hyperplane is given by $\nabla f^c - \nabla f^y$. At each iteration, the algorithm computes the minimal step size along this direction to reach the decision hyperplane. Since $f$ is not linear, the algorithm may need more than one iterations to cross the decision hyperplane.

We observe that if we set the loss function as $\mathcal{L}=f^c-f^y$ the gradient ascent direction matches the $\mathtt{DF}$ step directions $\nabla f^c - \nabla f^y$. We thus regard $\mathtt{DF}$ as a gradient ascent algorithm with each step size minimised to just induce a wrong prediction.

\paragraph{Projection and clipping}

The norm constraint $||\cdot||_2\leq\epsilon$ as well as RGB value constraint to $[0,255]$ must be enforced on the solution. \cite{Liu2016DelvingIT,kurakin2016adversarial} suggest applying projections after each iteration. We follow this practice. For BW images, we average the gradients for each RGB channel.

\subsection{\label{sec:our-aips}Our AIP methods}

\noindent
As we will demonstrate in \S \ref{sec:Comparison-Adversarial}, the above approaches are fragile to simple image processing techniques. We propose novel AIP approaches here, focusing on robustness.

\paragraph[.]{Gradient Ascent -- Maximal Among Non-GT ($\mathtt{GAMAN}$\protect\footnotemark)}
\refstepcounter{footnote}
\footnotetext{\emph{Gaman} is a Zen Buddhist term for \emph{endurance}.}
\addtocounter{footnote}{-1}

Even if the prediction label is changed by the AIP, this would not be robust if the perturbed input is still close to the decision boundary. DeepFool ($\mathtt{DF}$) is not expected to be robust, as it stops iterations as soon as the decision boundary is reached. On the other hand, $\mathtt{DF}$ guides the solution to the closest decision boundary; if we let $\mathtt{DF}$ iterate beyond the decision boundary with a fixed step size with fixed number of iterations, the solution is likely to proceed more deeply into the territory of the wrong label, improving robustness.

This motivates our $\mathtt{GAMAN}$ variant. Instead of the costly computation of $c$ at each iteration, we approximate $c\approx y^\star :=\underset{y^\prime\neq y}{\arg\min}f^{y^\prime}$, the most likely prediction among wrong labels. We set the loss function as $\mathcal{L}=f^{y^\star}-f^y$, and perform gradient ascent with a fixed step size $\gamma$ for $K$ iterations. This approach is similar but different from the impersonation AIPs previously considered \cite{Sharif16AdvML,Liu2016DelvingIT}, which drive the solution to a fixed impersonation target $\bar{y}$. In contrast, $y^{\star}$ may change during the iterations.

\paragraph{Vaccination against image processing}

The above methods maximise classification loss functions with respect to a fixed recogniser. For countering an AIP-neutralising image processing technique $n_j$, we consider including the image processing step in the loss function: $\mathcal{L}(n_j(x+t))$. Any first-order method considered above can be used, as long as $n_j$ is differentiable. If the processing function is random, we average the gradients from multiple samples. We refer to this technique as \emph{vaccination}. Note that this technique is complimentary to the above mentioned methods.

\paragraph{Selective AIPs}

We present another complimentary technique for generating AIPs targetted to a selected subset of recognisers. To avoid recognition from $\mathcal{M}$ while authorising $\mathcal{B}$ to recognise, we propose to maximise a mixed loss
\vspace{-0.5em}
\begin{equation}
\label{eq:adversarial-perturbation-multiple-recogniser}
\underset{k\in\mathcal{M}}{\sum}\lambda_k\mathcal{L}_k
-\underset{k^\prime\in\mathcal{B}}{\sum}\lambda_{k^\prime}\mathcal{L}_{k^\prime}
\vspace{-0.5em}
\end{equation}
with $\lambda_k,\lambda_{k^\prime}>0$.

\section{\label{sec:Experiments}Empirical Studies}

\noindent
We have set up a game theoretical framework to study the dynamics between the user $U$ and the recogniser $R$. In particular, previous adversarial image perturbation (AIP) techniques are studied, and new variants are proposed.

In this section, we present a case study of the framework on \emph{person recognition}. Before presenting the game theoretical analysis, we evaluate the performance of existing and newly proposed AIP techniques (\S \ref{sec:Comparison-Adversarial}), and the effectiveness of $R$'s image processing strategies $\Theta^r$ (\S \ref{sec:Robustness-Analysis}). The full game is introduced (\S \ref{sec:Game-Experiments}) after specifying $U$'s strategy space; we study this system in depth. Finally, we show results on the recogniser-selective AIPs (\S \ref{sec:Multi-Agent}).

\subsection{\label{sec:Dataset-and-Setup}Dataset and Experimental Setup}

\paragraph{Dataset}

We build our analysis upon the PIPA (People In Photo Albums) \cite{Zhang2015CvprPiper},
a large-scale dataset of social media photos crawled from Flickr. We use the $\mathtt{val}_1$ subset of PIPA, consisting of $4820$ instances of $366$ identities ($\mathtt{val}\text{-}\mathtt{original}$ $\mathtt{split}_1$ in \cite{Oh2015Iccv} terminology) as the test set. We assume that the user uploads cropped head images to social media; PIPA provides the GT head boxes.

\paragraph{Person recogniser}

The person recognition model $f$ is built on a state of the art framework \cite{Oh2015Iccv}. It first trains a convnet for the person recognition task on a large database of random identities; it then tunes the final classification layer to the test identities using about ten examples per identity. In our case, we have used the $\mathtt{val}_0$, which is of  the same size  and set of identities as as $\mathtt{val}_1$. While \cite{Oh2015Iccv} only considered AlexNet \cite{Krizhevsky2012Nips}, we also consider VGG \cite{Simonyan14vgg}, GoogleNet \cite{Szegedy15googlenet}, and ResNet152 \cite{He2016DeepRL}. They show better recognition rates (table \ref{tab:adv-performance}).

\paragraph{Evaluation}

We evaluate payoffs for $R$ in terms of the ratio of correctly identified instances in the test set. The payoff for $U$ is $1$ minus $R$'s payoff. In all the tables, $R$ is the column player and $U$ is the row player. For each column (row), $U$'s ($R$'s) optimal strategy is marked \ul{orange} (\hl{blue}).

\subsection{\label{sec:Comparison-Adversarial}Comparison of Perturbation Methods}

\begin{table}
\newbox\zerounderlined
\sbox\zerounderlined{\ul{0.0}}

\begin{centering}
\setlength\tabcolsep{0.3em}
\begin{tabular}{ccccccccc}
& & & Perturbation &  & AlexNet & VGG & Google & ResNet\tabularnewline
\vspace{-1em}
& & & &  &  &  &  & \tabularnewline
\cline{4-4} \cline{6-9} 
\vspace{-.9em}
& & & &  &  &  &  & \tabularnewline
\cline{4-4} \cline{6-9} 
\vspace{-1em}
& & & &  &  &  &  & \tabularnewline
& & &None &  & 83.8 & 86.1 & 87.8 &  \hl{91.1} \tabularnewline
\vspace{-1em}
& & & &  &  &  &  & \tabularnewline
\cline{1-2} \cline{4-4} \cline{6-9} 
\vspace{-1em}
& & & &  &  &  &  & \tabularnewline
\multirow{3}{*}{\rotatebox{90}{Image\hspace{0.5em}}} & \hspace{-.5em} \multirow{3}{*}{\rotatebox{90}{Proc.\hspace{0.5em}}}  &  & Noise &  & $\geq$83 & $\geq$85 & $\geq$87 &  \hl{$\geq$90}  \tabularnewline
& & & Blur &  & $\geq$82 & $\geq$85 & $\geq$86 & \hl{$\geq$90} \tabularnewline
& & & Eye Bar &  & $\geq$81 & $\geq$84 & $\geq$84 & \hl{$\geq$87} \tabularnewline
\vspace{-1em}
& & & &  &  &  &  & \tabularnewline
\cline{1-2} \cline{4-4} \cline{6-9} 
\vspace{-1em}
& & & &  &  &  &  & \tabularnewline
\multirow{2}{*}{\rotatebox{90}{1-Iter.\hspace{0.1em}}} & \hspace{-.5em} \multirow{2}{*}{\rotatebox{90}{AIP\hspace{0.5em}}} & & $\mathtt{FGS}$\cite{goodfellow2015iclr} &  & \hl{23.6} & 16.0 & 5.9 & 20.2\tabularnewline 
& & & $\mathtt{FGV}$\cite{Rozsa2016AdversarialDA} &  & 13.3 & 11.5 & 4.6 & \hl{20.0}\tabularnewline
\vspace{-1em}
& & & &  &  &  &  & \tabularnewline
\cline{1-2} \cline{4-4} \cline{6-9} 
\vspace{-1em}
& & & &  &  &  &  & \tabularnewline
\multirow{4}{*}{\rotatebox{90}{$K$-Iter.\hspace{0.5em}}} & \hspace{-.5em} \multirow{4}{*}{\rotatebox{90}{AIP\hspace{0.5em}}} & & $\mathtt{BI}$\cite{kurakin2016adversarial} &  & \hl{1.2} & 0.5 & \ul{0.0} & \ul{0.0} \tabularnewline
& & & $\mathtt{GA}$ &  & \hl{0.2} & \ul{0.0} & \ul{0.0} & \ul{0.0} \tabularnewline
\vspace{-1.2em} & & & &  &  &  &  & \tabularnewline
 \cline{4-4} \cline{6-9} 
\vspace{-1em} & & & &  &  &  &  & \tabularnewline
& & & $\mathtt{DF}$\cite{moosavi2016cvpr} &  & \hl{\mbox{\usebox\zerounderlined}} & \hl{\mbox{\usebox\zerounderlined}} & \hl{\mbox{\usebox\zerounderlined}} & \hl{\mbox{\usebox\zerounderlined}}\tabularnewline
\vspace{-1.2em} & & & &  &  &  &  & \tabularnewline
& & & $\mathtt{GAMAN}$ &  & \hl{\mbox{\usebox\zerounderlined}} & \hl{\mbox{\usebox\zerounderlined}} & \hl{\mbox{\usebox\zerounderlined}} & \hl{\mbox{\usebox\zerounderlined}}\tabularnewline
\vspace{-1.2em}
& & & &  &  &  &  & \tabularnewline
\cline{1-2} \cline{4-4} \cline{6-9} 
\end{tabular}
\par\end{centering}
\vspace{0em}

\caption{\label{tab:adv-performance}Recognition rates 
after image perturbation. In all methods, the perturbation is restricted to $||\cdot ||_2 \leq 1000$. For the baseline image processing perturbations, we only report lower bounds (denoted $\geq\cdot\,\,$).}
\end{table}

\paragraph{AIP parameters}

We set $\epsilon=1000$ in all our experiments, unless stated otherwise. For GoogleNet input $224\times 224$, this corresponds to 2\% of pixels perturbed by 1/256. For Gradient Ascent ($\mathtt{GA}$) and Basic Iterative ($\mathtt{BI}$) the step size $\gamma$ is set to $10^4$; for $\mathtt{GAMAN}$, $5\times 10^3$. We set the maximal number of iterations $K=100$, determined such that the norm reaches $\epsilon=1000$ in $K$ iterations for most test samples.

\paragraph{Baseline perturbation methods}

We consider three commonly used obfuscation types: noise, blur, and eye bar. Noise adds iid Gaussian noise of variance $\sigma^n$; blur performs convolution with a Gaussian kernel of size $\sigma^b$; eye bar puts a gray horizontal bar of thickness $\sigma^e$ on the upper $\frac{1}{3}$ location. They incur large $L_2$ distances ($>\negthinspace \negthinspace 1000$) from the original image even with small $\sigma^n$, $\sigma^b$, and $\sigma^e$. In table \ref{tab:adv-performance}, we report the \emph{lower bounds} on the recognition rates at $||\cdot||_2=1000$ by computing the rates at some $||\cdot||_2>1000$.

\paragraph{AIP performance}

We first evaluate all the considered AIP methods against all network variants. Table \ref{tab:adv-performance} shows the results. We observe that noise, blur, and eye bar have nearly no impact on the recognition performance for small $L_2$ perturbations. AIP variants show better obfuscation performances. Vanilla gradient overall gives better obfuscation than signed versions; on AlexNet Fast Gradient Vector ($\mathtt{FGV}$) reduces the recognition rate to 13.3, compared to 23.6 for Fast Gradient Sign ($\mathtt{FGS}$); the multi-iteration analogues show similar behaviours with Gradient Ascent ($\mathtt{GA}$) achieving 0.2 compared to 1.2 by Basic Iterative ($\mathtt{BI}$). Finally, we observe that the DeepFool ($\mathtt{DF}$) and $\mathtt{GAMAN}$ (\S \ref{sec:our-aips}) are very effective, pushing the recognition rates down to zero. 

\paragraph{Network performance}

Comparing architectures, we observe that AlexNet is surprisingly robust to AIPs compared to more recent architectures. GoogleNet, for example, performs better than Alexnet without AIPs (83.8 vs 87.8); when $\mathtt{FGS}$ is used, AlexNet performs 23.6 while GoogleNet performs 5.9. When multi-iteration AIPs are used, the architectural choice does not have a significant impact. We opt for GoogleNet in the next experiments; it is reasonably performant, while being much faster than ResNet.

\subsection{\label{sec:Robustness-Analysis}Robustness of AIPs}

\paragraph{Basic processing {$\mathsf{Proc}$}}

Even before $R$'s image processing strategies take place, the perturbed image needs to be (1) re-sized to the original image (from the network input sizes) and (2) quantised to integer values (\eg 24-bit true color). We denote the above two basic processing steps as $\mathsf{Proc}$.

\paragraph{Image processing strateges {$\Theta^r$}}

We fully specify $R$'s strategy space for our case study. Following Graese \etal \cite{graese2016icmla}, we consider $\Theta^r=\{\mathsf{Proc},\,\mathsf{T},\,\mathsf{N},\,\mathsf{B},\,\mathsf{C},\,\mathsf{TNBC}\}$. $\mathsf{Proc}$ is the basic processing described above, and all the other strategies are applied over $\mathsf{Proc}$. {$\mathsf{T}$} is translation by a random offset within $10\%$ of the image side lengths. {$\mathsf{N}$} adds iid Gaussian noise with variance $\sigma^2=10^2$. {$\mathsf{B}$} blurs with Gaussian kernel of width chosen from $\{1,3,5,7,9\}$ uniformly at random. {$\mathsf{C}$} crops with a random offset within $10\%$ of the image side lengths and re-sizes back to the original. For each strategy, the recogniser ensembles the scores from five random samples. We also consider the combination of all four ({$\mathsf{TNBC}$}). It runs the model four times on each processed image and once on the original; the scores are then averaged.

\begin{table}
\newbox\minimaxunderlined
\sbox\minimaxunderlined{\ul{22.2}}
\begin{centering}
\setlength\tabcolsep{0.4em}
\begin{tabular}{ccccccccc}
Perturb &  & $\emptyset$ & $\mathsf{Proc}$ & $\mathsf{T}$ & $\mathsf{N}$ & $\mathsf{B}$ & $\mathsf{C}$ & {\footnotesize $\mathsf{TNBC}$}\tabularnewline
\cline{1-1} \cline{3-9} 
\vspace{-1em}
 &  &  &  &  &  &  &  & \tabularnewline
\cline{1-1} \cline{3-9} 
\vspace{-.7em}
 &  &  &  &  &  &  &  & \tabularnewline
None &  & \hl{87.8} & \hl{87.8} & 87.6 & 64.0 & 81.2 & 85.4 & 87.3\tabularnewline
\cline{1-1} \cline{3-9} 
\vspace{-1em}
 &  &  &  &  &  &  &  & \tabularnewline
$\mathtt{BI}$\cite{kurakin2016adversarial} &  & \ul{0.0} & 8.3 & 15.8 & 16.8 & \hl{28.6} & 27.4 & 17.6 \tabularnewline
$\mathtt{GA}$ &  & \ul{0.0} & 8.6 & 13.2 & \ul{14.1} & \hl{28.4} & 23.7 & 16.4 \tabularnewline
\vspace{-1.2em}
 &  &  &  &  &  &  &  & \tabularnewline
\cline{1-1} \cline{3-9} 
\vspace{-1em}
 &  &  &  &  &  &  &  & \tabularnewline
$\mathtt{DF}$\cite{moosavi2016cvpr} &  & \ul{0.0} & 51.8 & 75.6 & 56.5 & 72.5 & \hl{76.9} & 75.5\tabularnewline
$\mathtt{GAMAN}$ &  & \ul{0.0} & \ul{4.0} & \ul{6.6} & 15.0 & \hl{\mbox{\usebox\minimaxunderlined}} & \ul{16.7} & \ul{9.9} \tabularnewline
\cline{1-1} \cline{3-9} 
\end{tabular}
\par\end{centering}
\vspace{0em}

\caption{\label{tab:adv-robustness}Robustness analysis of AIPs on GoogleNet. AIPs are restricted to
to $||\cdot ||_2 \leq 1000$. $\mathsf{Proc}$ indicates the re-sizing and quantisation needed to convert AIP outputs to image files. $(\mathsf{T},\mathsf{N},\mathsf{B},\mathsf{C})=$ (Translate, Noise, Blur, Crop).}
\end{table}

\paragraph{Robustness of AIPs}

Table \ref{tab:adv-robustness} shows the recognition rates for the GoogleNet when $R$'s processing strategies are present. While the multi-iteration AIPs induce zero recognition rates without any processing, $\mathsf{Proc}$ already exhibits powerful neutralisation effects: recognition rates for Gradient Ascent ($\mathtt{GA}$) and DeepFool ($\mathtt{DF}$) jump from zero to 8.6 and 51.8, respectively. The instability of $\mathtt{DF}$ is due to early stopping (\S \ref{sec:exising-aips}). The processing strategies by $R$ further increase recognition rates. Blurring $\mathsf{B}$ and cropping $\mathsf{C}$ strategies prove to be more harmful to AIPs than translation $\mathsf{T}$ and noise $\mathsf{N}$ in general. Comparing AIP-wise, we show that our novel variant $\mathtt{GAMAN}$ (\S \ref{sec:our-aips}) dominates other methods against all processing strategies but $\mathsf{N}$; $\mathtt{GA}$ performs better in that case, but only by a small amount (14.1 versus 15.0). Subsequent analyses are built on $\mathtt{GAMAN}$.

\begin{figure*}[t]
\begin{centering}
{\footnotesize{}}%
\begin{tabular}{ccccccc}
Original & & Blur & $\mathtt{GA}$ & $\mathtt{DF}\cite{moosavi2016cvpr}$ & $\mathtt{GAMAN}$ & $\mathtt{GAMAN}$  \tabularnewline
 $L_2=0$ &  & $L_2=4107$  &  $L_2=1000$  &  $L_2=119$  &  $L_2=1000$  & $L_2=2000$  \tabularnewline
\includegraphics[width=0.30\columnwidth]{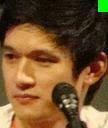} &
&
\includegraphics[width=0.30\columnwidth]{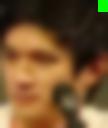} & \includegraphics[width=0.30\columnwidth]{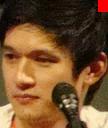} & \includegraphics[width=0.30\columnwidth]{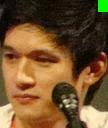} & \includegraphics[width=0.30\columnwidth]{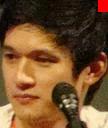} & \includegraphics[width=0.30\columnwidth]{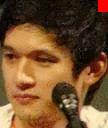}  \tabularnewline
 $L_2=0$ &  & $L_2=5666$  &  $L_2=1000$  &  $L_2=173$  &  $L_2=1000$  & $L_2=2000$  \tabularnewline
\includegraphics[width=0.30\columnwidth]{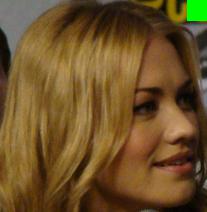} &
&
\includegraphics[width=0.30\columnwidth]{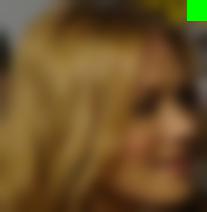} & \includegraphics[width=0.30\columnwidth]{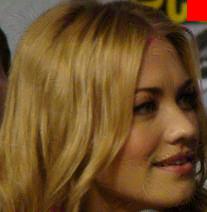} & \includegraphics[width=0.30\columnwidth]{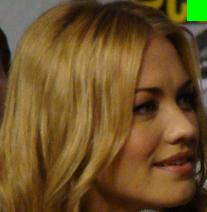} & \includegraphics[width=0.30\columnwidth]{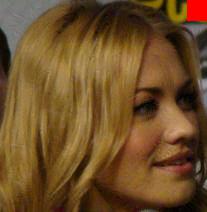} & \includegraphics[width=0.30\columnwidth]{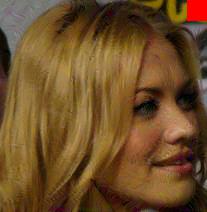}  \tabularnewline
\end{tabular}
\par\end{centering}{\footnotesize \par}
\vspace{0em}

\caption{\label{fig:example-adversarial-perturbation}Perturbed images after $\mathsf{Proc}$ and the corresponding predictions (green for correct, red for wrong). $\mathtt{GA}$ and $\mathtt{GAMAN}$ reliably confuse the classifier at almost no cost on the aesthetics. At $L_2=2000$, $\mathtt{GAMAN}$ does show small artifacts.}
\end{figure*}

\paragraph{Qualitative}

Qualitative examples of the methods are shown in figure \ref{fig:example-adversarial-perturbation}. The images and the prediction results are after $\mathsf{Proc}$. $\mathtt{GA}$ and $\mathtt{GAMAN}$ reliably induces misidentification without sacrificing aesthetics compared to blurring.

\subsection{\label{sec:Game-Experiments}User-Recogniser Games}

\paragraph{Vaccination strategies {$\Theta^u$}}

In response to the processing strategies by the recogniser $R$, the user $U$ may vaccinate the AIP against expected processing types (\S \ref{sec:our-aips}). We consider six variants $\Theta^u=\{\mathtt{GAMAN},\mathtt{/T},\mathtt{/N},\mathtt{/B},\mathtt{/C},\mathtt{/TNBC}\}$. We use slash $\mathtt{/}$ to indicate vaccination on $\mathtt{GAMAN}$. For $\mathtt{/T},\mathtt{/N},\mathtt{/B},\mathtt{/C}$, gradients from 5 random function samples are averaged at each iteration. The combination strategy $\mathtt{/TNBC}$ averages 4 gradients from individual methods and 1 original gradient, resulting in the same number of gradient computations for all vaccination variants.

\paragraph[?]{Is vaccination helpful}

Table \ref{tab:adv-counter} shows the recognition rates of GoogleNet for combinations of discussed processing and vaccination strategies. We observe indeed that each vaccination type makes the vanilla AIP $\mathtt{GAMAN}$ more robust against the respective processing type: for $\mathsf{B}$ the rate drops from 22.2 to 5.8. $\mathtt{/B}$ is the most effective strategy for $U$ against all processing strategies except for $\mathsf{N}$. For $\mathsf{N}$, the corresponding vaccination $\mathtt{/N}$ yields the best payoff for $U$. We conjecture this is because the noise $\mathsf{N}$ results in high frequency patterns while the others smooth the output. We observe, finally, that the combined vaccination $\mathtt{/TNBC}$ cannot prepare AIP against all processing types most effectively; given a budget on the number of gradient computations, it is hard to be good at everything.

\begin{table}
\begin{centering}
\setlength\tabcolsep{0.5em}
\begin{tabular}{cccccccc}
 &  & \multicolumn{6}{c}{Recogniser $\Theta^r$}\tabularnewline
\vspace{-1em}
 &  &  &  &  &  &  & \tabularnewline
\cline{3-8} 
\vspace{-1em}
 &  &  &  &  &  &  & \tabularnewline
User $\Theta^u$ &  & $\mathsf{Proc}$ & $\mathsf{T}$ & $\mathsf{N}$ & $\mathsf{B}$ & $\mathsf{C}$ & {\footnotesize $\mathsf{TNBC}$}\tabularnewline
\vspace{-1em}
 &  &  &  &  &  &  & \tabularnewline
\cline{1-1} \cline{3-8} 
\vspace{-1em}
 &  &  &  &  &  &  & \tabularnewline
\cline{1-1} \cline{3-8} 
\vspace{-1em}
 &  &  &  &  &  &  & \tabularnewline
$\mathtt{GAMAN}$ &  & 4.0 & 6.6 & 15.0 & \hl{22.2} & 16.7 & 9.9 \tabularnewline
$\mathtt{/T}$ &  & 2.5 & 2.3 & 11.6 & \hl{18.5} & 7.2 & 4.9 \tabularnewline
$\mathtt{/N}$ &  & 5.8 & 7.6 & \ul{4.6} & \hl{23.6} & 16.6 & 9.1 \tabularnewline
$\mathtt{/B}$ &  & \ul{0.4} & \ul{0.8} & \hl{8.6} & \ul{5.8} & \ul{3.1} & \ul{1.4} \tabularnewline
$\mathtt{/C}$ &  & 2.6 & 2.2 & 11.8 & \hl{18.1} & 3.4 & 4.3 \tabularnewline
$\mathtt{/TNBC}$ &  & 0.7 & 0.9 & 5.2 & \hl{9.5} & 3.2 & 2.0 \tabularnewline
\vspace{-1em}
 &  &  &  &  &  &  &\tabularnewline
\cline{1-1} \cline{3-8} 
\end{tabular}
\par\end{centering}
\vspace{0em}

\caption{\label{tab:adv-counter}Recogniser's payoff table $p_{ij}$, $i\in \Theta^u$ and $j\in\Theta^r$. The user's payoff is given by $100-p_{ij}$.}
\end{table}

\paragraph{Optimal deterministic strategy}

We can regard table \ref{tab:adv-counter} as the payoff table $p_{ij}$ for $R$ for strategies $i\in\Theta^u$ and $j\in\Theta^r$. Let's first assume that the players only choose fixed strategies. Then, solving equation \ref{eq:minimax-optimisation} with determinism constraints $\theta^u_i,\theta^r_j\in\{0,1\}$ yields $U$'s optimal strategy as $\mathtt{/B}$ with a privacy guarantee of at most 8.6 recognition rate. 

\paragraph{Optimal random strategy}

Game theory suggests that it is sometimes better to randomise strategies. Solving equation \ref{eq:minimax-optimisation} without the integral constraints yield the optimal solutions for $U$ and $R$ as $\theta^{u\star}=\left(\mathtt{/B}:61\%,\mathtt{/TNBC}:39\%\right)$ and $\theta^{r\star}=\left(\mathsf{N}:52\%,\mathsf{B}:48\%\right)$, respectively. Playing $\theta^{u\star}$ guarantees $U$ to allow at most 7.3 recognition rate, an improved privacy guarantee than the deterministic case, 8.6.

\paragraph{Knowledge on $R$'s strategy}

As discussed in \S \ref{sec:Game-Theory}, having knowledge on $R$'s strategy can improve the payoff bound for $U$, if $R$ does not play the optimal strategy. Let us consider two possible non-optimal strategies played by $R$. (1) If $R$ commits to $\mathsf{B}$, $U$'s optimal strategy is the minimal row in the column $\mathsf{B}$: $\mathtt{/B}$, with recognition rate 5.8. (2) If $R$ randomises uniformly over $\Theta^r$, $U$'s optimal strategy is the minimal row over the column average: $\mathtt{/B}$ with recognition rate 3.4. In both cases, $U$ enjoys lower recognition rates.

\paragraph{Limited knowledge on {$\Theta^r$}}

Assume that $U$ is not aware of all possible technologies that $R$ has at hand. For example, the strategy $\mathsf{N}$ is not known to $U$. Then, $U$'s apparent optimal solution is $\left(\texttt{/B}:100\%\right)$, which she thinks will guarantee her at most 5.8 recognition rate. $R$ can then attack $U$ with $\mathsf{N}$, incurring 8.6 recognition rate. Limited knowledge on the opponent's strategy space does hurt.

\subsection{\label{sec:Multi-Agent}Selective AIPs}

\begin{table}
\begin{centering}
\setlength\tabcolsep{0.3em}
\begin{tabular}{ccccccccc}
 
 \multicolumn{3}{c}{Setup}  &   & \multicolumn{2}{c}{$\mathcal{M}$ averaged} & &  \multicolumn{2}{c}{$\mathcal{B}$ averaged} \tabularnewline
 
$\mathcal{M}$ & $\mathcal{B}$ & $L_2$ & & {\small w/o AIP} & {\small w/ AIP} &  & {\small w/o AIP} & {\small w/ AIP} \tabularnewline

\cline{1-3} \cline{5-6} \cline{8-9} 
\vspace{-1em}
 &  &  &  & &  & \tabularnewline
\cline{1-3} \cline{5-6} \cline{8-9} 
\vspace{-1em}
 &  &  &  & &  & \tabularnewline
$\left\{ \text{G}\right\} $ & $\emptyset$ & 1000 &  & 87.8 & 4.0 & & - & -\tabularnewline
\vspace{-1em}
 &  &  &  & &  & \tabularnewline
$\left\{ \text{G}\right\} $ & $\left\{ \text{A}\right\} $ & 1000 &  & 
87.8 & 8.7 &  & 83.8 & 97.9\tabularnewline
\vspace{-1em}
 &  &  & & &  & \tabularnewline
$\left\{ \text{A,R}\right\} $ & $\left\{ \text{V,G}\right\} $ & 1000 &  & 87.4 & 17.7 &  & 87.0 & 97.7 \tabularnewline
\vspace{-1em}
 &  &  & & &  & \tabularnewline
$\left\{ \text{A,R}\right\} $ & $\left\{ \text{V,G}\right\} $ & 2000 &  & 87.4 & 3.8 &  & 87.0 & 97.8\tabularnewline
\vspace{-1em}
 &  &  & & &  & \tabularnewline
\cline{1-3} \cline{5-6} \cline{8-9} 

\end{tabular}
\par\end{centering}
\vspace{0em}

\caption{\label{tab:adv-multiagent}Selective AIPs. AIPs are crafted to confuse $\mathcal{M}$ leaving $\mathcal{B}$ intact. [A,V,G,R] = [AlexNet, VGG, GoogleNet, ResNet152]. $\mathtt{GAMAN}$ has been used in all experiments. Reported performances are after $\mathsf{Proc}$.}
\end{table}

\noindent
We assume that $U$ wants to avoid identification by a set of malicious recognisers $\mathcal{M}$, while authorising identification by benign ones $\mathcal{B}$. We set up the experiments in table \ref{tab:adv-multiagent}. We include the $\mathtt{GAMAN}$ performance on GoogleNet as a baseline (first row). We solve equation \ref{eq:adversarial-perturbation-multiple-recogniser} with $\lambda_k=1$ for all $k\in\mathcal{M}\cup\mathcal{B}$ to generate selective AIPs.

When $\mathcal{M}=\{\text{GoogleNet}\}$ and $\mathcal{B}=\{\text{AlexNet}\}$, the generated AIP incurs mere 8.7 identification for $\mathcal{M}$ (after $\mathsf{Proc}$), while allowing $\mathcal{B}$ to identify 97.9 percent. We thus confirm the selectivity. However, this comes at the cost of increased recognition rate for $\mathcal{M}$ (8.7), compared to when AIP only had to confuse $\mathcal{M}$ (4.0).

We also consider the multi-$\mathcal{M}$, multi-$\mathcal{B}$ case given by $\mathcal{M}=\{\text{AlexNet, ResNet}\}$ and $\mathcal{B}=\{\text{VGG, GoogleNet}\}$. The average performance is 17.7 for $\mathcal{M}$, and 97.7 for $\mathcal{B}$, post $\mathsf{Proc}$. Selectivity thus works for multiple models, but again the recognition rates for $\mathcal{M}$ are quite high (17.7). We remark that by increasing the budget on perturbation size from $1000$ to $2000$, we can still attain a lower rate: 3.8.

The existence of selective AIPs is not only of practical but also of theoretical interest. They show that the space of AIPs is diverse enough to accommodate patterns that simultaneously hamper and assist recognition.

\section{\label{sec:Conclusion}Discussion \& Conclusion}

\paragraph{Game theoretical approach}

Game theory is a tool for wading through uncertainties in players' choices, providing payoff guarantees independent of the opponent's strategies. Game theory also suggests that if there is no single technology which best copes with all possible adversarial technologies, it is better to randomise existing techniques.

As discussed in \S \ref{sec:limitations}, the game theoretical framework introduced in this paper can be extended to other setups, where less resource constraints are placed on each player. This paper serves as a first step towards the promising research direction of analysing the user-recogniser dynamics.

\paragraph{Conclusion}

In this work, we have constructed a game theoretical framework to study a system with two players, user $U$ and recogniser $R$, with antagonistic goals (dis-/enable recognition). We have examined existing and new adversarial image perturbation (AIP) techniques for $U$. As a case study of the framework, we have presented a game theoretical analysis of the privacy guarantees for a social media user, assuming strategy spaces that include the state of the art AIPs and person recognition techniques.

\paragraph{Acknowledgement}

This research was supported by the German Research Foundation (DFG CRC 1223). We thank Dr Yun Kuen Cheung for exciting discussions on the Game Theory and comments on the manuscript. We also thank Dr Mykhaylo Andriluka and Tribhuvanesh Orekondy for helpful comments on the paper.

\FloatBarrier


\bibliographystyle{ieee}
\bibliography{iccv_2017_adversarial_privacy}


\newpage
\newpage
\pagebreak
\clearpage

\appendix
\part*{Supplementary Materials}

\section{\label{sec:Contents}Contents}

\noindent
The supplementary materials contain auxiliary experiments for the empirical analyses in the main paper. In particular, we include:

\begin{itemize}
\item Score loss for adversarial image perturbation (AIP).
\item AIP performance at different $L_2$ norms.
\item Experiments for the non-GoogleNet architectures.
\item More qualitative results.
\end{itemize}

\noindent
As in the main paper, we mark the optimal entry in each column  (row) for the user (recogniser) with \ul{orange} (\hl{blue}).

\section{\label{sec:more-aips}Score Loss for AIPs}

\noindent
In the main paper, we have reviewed variants of AIPs according to the loss functions and the optimisation algorithms. Algorithms $\mathtt{FGV}$, $\mathtt{FGS}$, $\mathtt{BI}$, and $\mathtt{GA}$ use the softmax-log loss $-\log \hat{f}^y$. The DeepFool ($\mathtt{DF}$) and our $\mathtt{GAMAN}$ variants use the difference of two scores (\eg $f^{y^\star}-f^y$). This section includes an auxiliary analysis for the effect of the loss type: softmax-log loss $-\log \hat{f}^y$ versus score loss $-f^y$. We denote the score loss analogues with the suffix $\text{-}\mathtt{S}$ (\eg $\mathtt{FGS}\text{-}\mathtt{S}$). We also include $\mathtt{FGMAN}$ (Fast Gradient -- Maximal Among Non-GT), the single iteration analogue of $\mathtt{GAMAN}$, for completeness. See table \ref{tab:supp-table-aip-losses} for a summary.

The corresponding empirical performances are shown in table \ref{tab:supp-table-aip-accs} and \ref{tab:supp-multiarch-robustness}. Since single-iteration AIPs are significantly outperformed by the multi-iteration AIPs, we have focused on the latter in the main paper, and so do we here. In table \ref{tab:supp-table-aip-accs}, we observe that the choice of the loss function does not make much difference. Table \ref{tab:supp-multiarch-robustness} further supports this view against image processing techniques, although the softmax-log loss does perform marginally better.

\begin{table}
\begin{centering}
\begin{tabular}{ccccc}
\multirow{2}{*}{Variants} &  & \multirow{2}{*}{Loss $\mathcal{L}$} & Stopping  & \multirow{2}{*}{Step size}\tabularnewline
 &  &  & condition & \tabularnewline
\vspace{-1em}
 &  &  &  & \tabularnewline
\cline{1-1} \cline{3-5} 
\vspace{-1em}
 &  &  &  & \tabularnewline
\cline{1-1} \cline{3-5} 
\vspace{-1em}
 &  &  &  & \tabularnewline
$\mathtt{FGS}$\cite{goodfellow2015iclr} &  & $-\log \hat{f}^{y}$ & 1 iteration & Fixed\tabularnewline
\vspace{-1em}
 &  &  &  & \tabularnewline
$\mathtt{FGV}$\cite{Rozsa2016AdversarialDA} &  & $-\log \hat{f}^{y}$ & 1 iteration & Fixed\tabularnewline
\vspace{-1em}
 &  &  &  & \tabularnewline
\cline{1-1} \cline{3-5} 
\vspace{-1em}
 &  &  &  & \tabularnewline
\cellcolor{Gray}
$\mathtt{FGS}\text{-}\mathtt{S}$&  & $-f^{y}$ & 1 iteration & Fixed\tabularnewline
\vspace{-1em}
\cellcolor{Gray}
 &  &  &  & \tabularnewline
 \cellcolor{Gray}
$\mathtt{FGV}\text{-}\mathtt{S}$ &  & $-f^{y}$ & 1 iteration & Fixed\tabularnewline
\vspace{-1em}
 &  &  &  & \tabularnewline
\cline{1-1} \cline{3-5} 
\vspace{-1em}
 &  &  &  & \tabularnewline
 \cellcolor{Gray}
$\mathtt{FGMAN}$ &  & $f^{y^{\star}}-f^{y}$ & 1 iteration & Fixed\tabularnewline
\vspace{-1em}
 &  &  &  & \tabularnewline
\cline{1-1} \cline{3-5} 
\vspace{-1em}
 &  &  &  & \tabularnewline
$\mathtt{BI}$\cite{kurakin2016adversarial} &  & $-\log \hat{f}^y$ & $K$ iterations & Fixed\tabularnewline
\vspace{-1em}
 &  &  &  & \tabularnewline
$\mathtt{GA}$ &  & $-\log \hat{f}^y$ & $K$ iterations & Fixed\tabularnewline
\vspace{-1em}
 &  &  &  & \tabularnewline
\cline{1-1} \cline{3-5} 
\vspace{-1em}
 &  &  &  & \tabularnewline
 \cellcolor{Gray}
$\mathtt{BI}\text{-}\mathtt{S}$ &  & $-f^{{y}}$ & $K$ iterations & Fixed\tabularnewline
\vspace{-1em}
\cellcolor{Gray}
 &  &  &  & \tabularnewline
 \cellcolor{Gray}
$\mathtt{GA}\text{-}\mathtt{S}$ &  & $-f^{{y}}$ & $K$ iterations & Fixed\tabularnewline
\vspace{-1em}
 &  &  &  & \tabularnewline
\cline{1-1} \cline{3-5} 
\vspace{-1em}
 &  &  &  & \tabularnewline
 $\mathtt{DF}$\cite{moosavi2016cvpr} &  & $f^{y^{c}}-f^{{y}}$ & $K$ it.$\vee$ fooled & Adaptive\tabularnewline
\vspace{-1em}
 &  &  &  & \tabularnewline
$\mathtt{GAMAN}$ &  & $f^{y^{\star}}-f^{{y}}$ & $K$ iterations & Fixed\tabularnewline
\vspace{-1em}
 &  &  &  & \tabularnewline
\cline{1-1} \cline{3-5} 
\end{tabular}
\par\end{centering}
\vspace{0em}

\caption{\label{tab:supp-table-aip-losses}Extended version of table \ref{tab:review-adversarial-perturbation} in the main paper; additional methods are denoted as gray cells. $f^{y^\prime}$ is the model score for class $y^\prime$, and $\hat{f}$ denotes the softmax output of $f$. $y$ is the ground truth label, and $y^{\star}$ is the most likely label among wrong ones. $y^{c}$ is the label with the closest linearised decision boundary. $\tilde{y}$ is the least likely label.}
\end{table}

\begin{table}
\newbox\zerounderlined
\sbox\zerounderlined{\ul{0.0}}

\begin{centering}
\setlength\tabcolsep{0.3em}
\begin{tabular}{ccccccccc}
& & & Perturbation &  & AlexNet & VGG & Google & ResNet\tabularnewline
\vspace{-1em}
& & & &  &  &  &  & \tabularnewline
\cline{4-4} \cline{6-9} 
\vspace{-.9em}
& & & &  &  &  &  & \tabularnewline
\cline{4-4} \cline{6-9} 
\vspace{-1em}
& & & &  &  &  &  & \tabularnewline
& & &None &  & 83.8 & 86.1 & 87.8 &  \hl{91.1} \tabularnewline
\vspace{-1em}
& & & &  &  &  &  & \tabularnewline
\cline{1-2} \cline{4-4} \cline{6-9} 
\vspace{-1em}
& & & &  &  &  &  & \tabularnewline
\multirow{3}{*}{\rotatebox{90}{Image\hspace{0.5em}}} & \hspace{-.5em} \multirow{3}{*}{\rotatebox{90}{Proc.\hspace{0.5em}}}  &  & Noise &  & $\geq$83 & $\geq$85 & $\geq$87 &  \hl{$\geq$90}  \tabularnewline
& & & Blur &  & $\geq$82 & $\geq$85 & $\geq$86 & \hl{$\geq$90} \tabularnewline
& & & Eye Bar &  & $\geq$81 & $\geq$84 & $\geq$84 & \hl{$\geq$87} \tabularnewline
\vspace{-1em}
& & & &  &  &  &  & \tabularnewline
\cline{1-2} \cline{4-4} \cline{6-9} 
\vspace{-1em}
& & & &  &  &  &  & \tabularnewline
\multirow{5}{*}{\rotatebox{90}{1-Iter.\hspace{0.1em}}} & \hspace{-.5em} \multirow{5}{*}{\rotatebox{90}{AIP\hspace{0.5em}}} & & $\mathtt{FGS}$\cite{goodfellow2015iclr} &  & \hl{23.6} & 16.0 & 5.9 & 20.2\tabularnewline 
& & & $\mathtt{FGV}$\cite{Rozsa2016AdversarialDA} &  & 13.3 & 11.5 & 4.6 & \hl{20.0}\tabularnewline
\vspace{-1.2em} & & & &  &  &  &  & \tabularnewline
 \cline{4-4} \cline{6-9} 
\vspace{-1em} & & & &  &  &  &  & \tabularnewline
& & & \cellcolor{Gray}$\mathtt{FGS}\text{-}\mathtt{S}$ &  & \hl{27.8} & 6.2 & 1.0 & 4.3\tabularnewline
& & & \cellcolor{Gray}$\mathtt{FGV}\text{-}\mathtt{S}$ &  & \hl{21.0} & 5.5 & 3.5 & 8.0\tabularnewline 
\vspace{-1.2em} & & & &  &  &  &  & \tabularnewline
 \cline{4-4} \cline{6-9} 
\vspace{-1em} & & & &  &  &  &  & \tabularnewline
& & & \cellcolor{Gray}$\mathtt{FGMAN}$ &  & 4.4 & 3.9 & 2.8 & \hl{11.5} \tabularnewline
\vspace{-1em}
& & & &  &  &  &  & \tabularnewline
\cline{1-2} \cline{4-4} \cline{6-9} 
\vspace{-1em}
& & & &  &  &  &  & \tabularnewline
\multirow{6}{*}{\rotatebox{90}{$K$-Iter.\hspace{0.5em}}} & \hspace{-.5em} \multirow{6}{*}{\rotatebox{90}{AIP\hspace{0.5em}}} & & $\mathtt{BI}$\cite{kurakin2016adversarial} &  & \hl{1.2} & 0.5 & \ul{0.0} & \ul{0.0} \tabularnewline
& & & $\mathtt{GA}$ &  & \hl{0.2} & \ul{0.0} & \ul{0.0} & \ul{0.0} \tabularnewline
\vspace{-1.2em} & & & &  &  &  &  & \tabularnewline
 \cline{4-4} \cline{6-9} 
\vspace{-1em} & & & &  &  &  &  & \tabularnewline
& & & \cellcolor{Gray}$\mathtt{BI}\text{-}\mathtt{S}$ &  & \hl{1.2} & 0.3 & \ul{0.0} & \ul{0.0} \tabularnewline
\vspace{-1.2em} & & & &  &  &  &  & \tabularnewline
& & & \cellcolor{Gray}$\mathtt{GA}\text{-}\mathtt{S}$ &  & \hl{0.2} & \ul{0.0} & \ul{0.0} & \ul{0.0} \tabularnewline
\vspace{-1.2em} & & & &  &  &  &  & \tabularnewline
 \cline{4-4} \cline{6-9} 
\vspace{-1em} & & & &  &  &  &  & \tabularnewline
& & & $\mathtt{DF}$\cite{moosavi2016cvpr} &  & \hl{\mbox{\usebox\zerounderlined}} & \hl{\mbox{\usebox\zerounderlined}} & \hl{\mbox{\usebox\zerounderlined}} & \hl{\mbox{\usebox\zerounderlined}}\tabularnewline
\vspace{-1.2em} & & & &  &  &  &  & \tabularnewline
& & & $\mathtt{GAMAN}$ &  & \hl{\mbox{\usebox\zerounderlined}} & \hl{\mbox{\usebox\zerounderlined}} & \hl{\mbox{\usebox\zerounderlined}} & \hl{\mbox{\usebox\zerounderlined}}\tabularnewline
\vspace{-1.2em}
& & & &  &  &  &  & \tabularnewline
\cline{1-2} \cline{4-4} \cline{6-9} 
\end{tabular}
\par\end{centering}
\vspace{0em}

\caption{\label{tab:supp-table-aip-accs}Extended version of table \ref{tab:adv-performance} in the main paper; new entries are denoted as gray cells. Recognition rates 
after image perturbation. In all methods, the perturbation is restricted to $||\cdot ||_2 \leq 1000$. For the baseline image processing perturbations, we only report lower bounds (denoted $\geq\cdot\,\,$).}
\end{table}

\section{\label{sec:L2-perf}AIP Performance at Different $L_2$ Norms}

\begin{figure}[t]
\includegraphics[width=0.9\columnwidth]{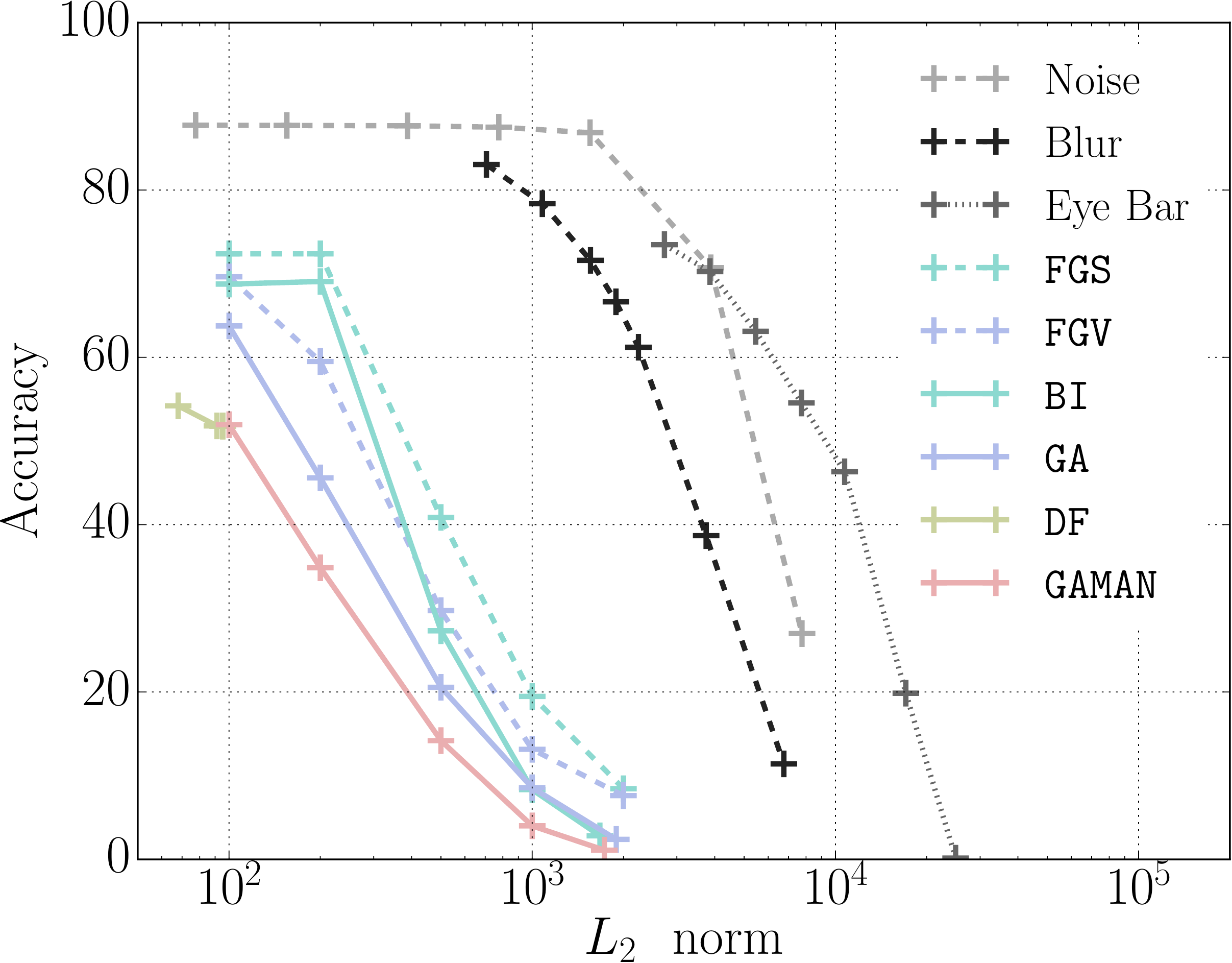}
\vspace{0em}

\caption{\label{fig:adv-plot-l2-acc}GoogleNet accuracy after various perturbations methods at different $L_{2}$ norms. All results are after $\mathsf{Proc}$.}
\end{figure}

\noindent
In the main paper, we have used the $L_2$ norm constraint $\epsilon=1000$ as the default choice. In this section, we examine the behaviour of AIP performance at varying $\epsilon$ values. 

See figure \ref{fig:adv-plot-l2-acc} for the plot. The performances are post-$\mathsf{Proc}$ (\S \ref{sec:Robustness-Analysis}). We fix the step size to $\gamma=10^4$ ($5\times 10^3$ for $\mathtt{GAMAN}$), and the maximal number of iterations to $K=100$; we choose the norm constraint $\epsilon$ from $\{100,200,500,1000,2000\}$. The norm of the resulting AIP is upper bounded by $\epsilon$, but may not necessarily be exactly $\epsilon$. The average norm across the test set is plotted. 

We observe that the AIP variants are much more effective than Noise, Blur, or Eye Bar, achieving the same degree of obfuscation at $1\negthinspace\sim\negthinspace 2$ orders of magnitude smaller perturbations. At the same norm level, the multi-iteration variants ($\mathtt{BI}$,$\mathtt{GA}$) are more effective than the single-iteration analogues ($\mathtt{FGS}$,$\mathtt{FGV}$). Taking gradient signs decreases the obfuscation performance at small $L_2$ norms ($\leq 1000$), but they converge to a similar performance at $\epsilon=2000$. DeepFool ($\mathtt{DF}$) outputs have small norms $\leq 100$ due to early stopping. Our variant $\mathtt{GAMAN}$ performs best across all norm levels, achieving nearly zero recognition at $\epsilon=2000$.

\section{\label{sec:more-architectures}Non-GoogleNet Experiments}

\noindent
In the main paper, we have focused on the GoogleNet results for the AIP robustness analysis and the game theoretic studies (table \ref{tab:adv-robustness} and \ref{tab:adv-counter}). We extend the experiments to AlexNet, VGG, and ResNet152.

\subsection{\label{sec:more-robustness}Robustness Analysis}

\noindent
See table \ref{tab:supp-multiarch-robustness} for the robustness analyses for all four networks. We confirm here again that $\mathtt{GAMAN}$ shows overall best robustness, across image processing techniques ($\mathsf{Proc}$, $\mathsf{T}$, $\mathsf{N}$, $\mathsf{B}$, $\mathsf{C}$, and $\mathsf{TNBC}$), across architectures. For AlexNet and ResNet, cropping ($\mathtt{C}$) is the most powerful neutralisation, while for VGG and GoogleNet blurring ($\mathtt{B}$) is. We observe that the effects are particularly strong for ResNet; $\mathtt{C}$ boosts the performance from 0.0 to 31.8 against $\mathtt{GAMAN}$.

\subsection{\label{sec:more-game}Game Analysis for Various Networks}

\noindent
See table \ref{tab:supp-multiarch-game} for the payoff tables for all four networks. We summarise the optimal user strategy $\theta^{u\star}$ and the corresponding guarantee on the recognition rate in table \ref{tab:supp-opt}. Note that against all but AlexNet architecture, the optimal strategy $\theta^{u\star}$ is given as a mixture of $/\mathtt{B}$ and $/\mathtt{TNBC}$.

\section{\label{sec:qualitative}Additional Qualitative Results}

\begin{figure*}[t]
\begin{centering}
{\footnotesize{}}%
\begin{tabular}{cccccccc}
Original & & Blur & $\mathtt{GA}$ & $\mathtt{DF}\cite{moosavi2016cvpr}$ & $\mathtt{GAMAN}$ & $\mathtt{GAMAN}$ & $\mathtt{GAMAN}$  \tabularnewline
 $L_2=0$ &  & $L_2=7425$  &  $L_2=1000$  &  $L_2=0$  &  $L_2=1000$  & $L_2=2000$  & $L_2=3000$  \tabularnewline
\includegraphics[width=0.25\columnwidth]{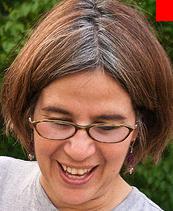} & &
\includegraphics[width=0.25\columnwidth]{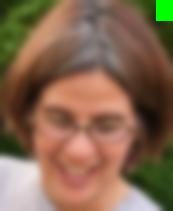} & 
\includegraphics[width=0.25\columnwidth]{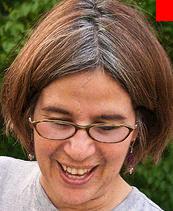} & 
\includegraphics[width=0.25\columnwidth]{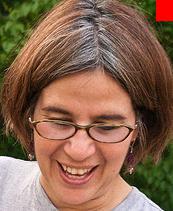} & 
\includegraphics[width=0.25\columnwidth]{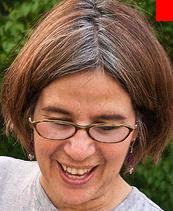} & 
\includegraphics[width=0.25\columnwidth]{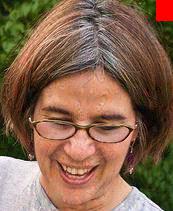} & 
\includegraphics[width=0.25\columnwidth]{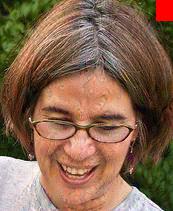} \tabularnewline
 & & & 
\includegraphics[width=0.25\columnwidth]{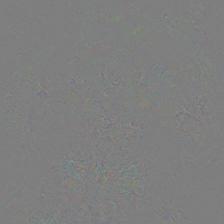} & 
\includegraphics[width=0.25\columnwidth]{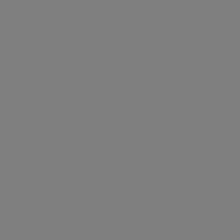} & 
\includegraphics[width=0.25\columnwidth]{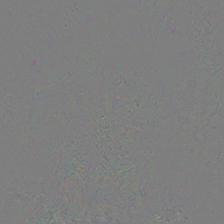} & 
\includegraphics[width=0.25\columnwidth]{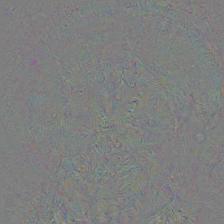} & 
\includegraphics[width=0.25\columnwidth]{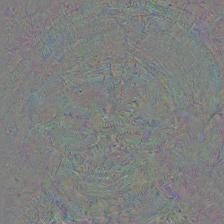} \tabularnewline
\\
 $L_2=0$ &  & $L_2=1865$  &  $L_2=1000$  &  $L_2=51$  &  $L_2=1000$  & $L_2=2000$  & $L_2=3000$  \tabularnewline
\includegraphics[width=0.25\columnwidth]{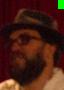} & &
\includegraphics[width=0.25\columnwidth]{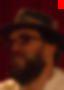} & 
\includegraphics[width=0.25\columnwidth]{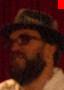} & 
\includegraphics[width=0.25\columnwidth]{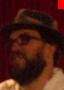} & 
\includegraphics[width=0.25\columnwidth]{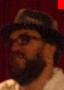} & 
\includegraphics[width=0.25\columnwidth]{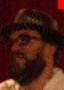} & 
\includegraphics[width=0.25\columnwidth]{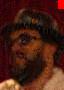} \tabularnewline
 & & & 
\includegraphics[width=0.25\columnwidth]{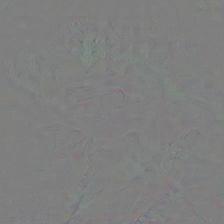} & 
\includegraphics[width=0.25\columnwidth]{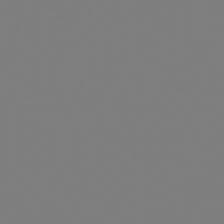} & 
\includegraphics[width=0.25\columnwidth]{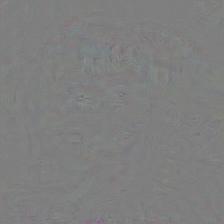} & 
\includegraphics[width=0.25\columnwidth]{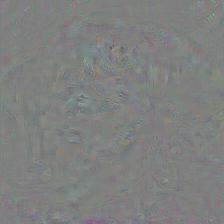} & 
\includegraphics[width=0.25\columnwidth]{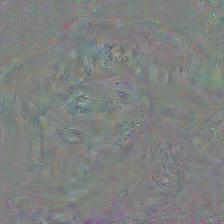} \tabularnewline
\\
 $L_2=0$ &  & $L_2=4067$  &  $L_2=1000$  &  $L_2=12$  &  $L_2=1000$  & $L_2=2000$  & $L_2=3000$  \tabularnewline
\includegraphics[width=0.25\columnwidth]{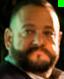} & &
\includegraphics[width=0.25\columnwidth]{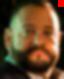} & 
\includegraphics[width=0.25\columnwidth]{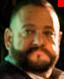} & 
\includegraphics[width=0.25\columnwidth]{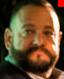} & 
\includegraphics[width=0.25\columnwidth]{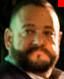} & 
\includegraphics[width=0.25\columnwidth]{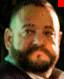} & 
\includegraphics[width=0.25\columnwidth]{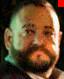} \tabularnewline
 & & & 
\includegraphics[width=0.25\columnwidth]{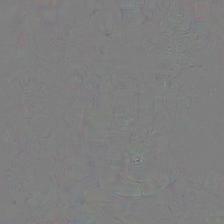} & 
\includegraphics[width=0.25\columnwidth]{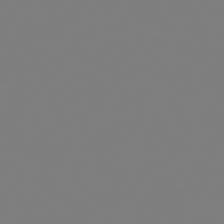} & 
\includegraphics[width=0.25\columnwidth]{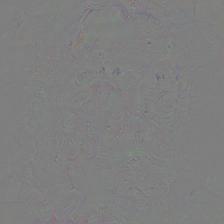} & 
\includegraphics[width=0.25\columnwidth]{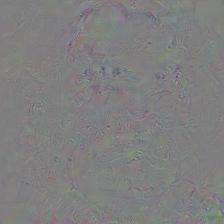} & 
\includegraphics[width=0.25\columnwidth]{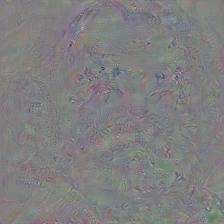} \tabularnewline

\end{tabular}
\par\end{centering}{\footnotesize \par}
\vspace{0em}

\caption{\label{fig:example1}Randomly chosen perturbed images after $\mathsf{Proc}$ and the corresponding GoogleNet predictions (green for correct, red for wrong). Perturbations are visualised with gray background. $\mathtt{GA}$ and $\mathtt{GAMAN}$ reliably confuse the classifier at almost no cost on the aesthetics. As the $L_2$ norm increases, artifacts become more visible. Perturbations may be too small to be visible when printed; zoom in in electronic version for better visibility.}
\end{figure*}

\begin{figure*}[t]
\begin{centering}
{\footnotesize{}}%
\begin{tabular}{cccccccc}
Original & & Blur & $\mathtt{GA}$ & $\mathtt{DF}\cite{moosavi2016cvpr}$ & $\mathtt{GAMAN}$ & $\mathtt{GAMAN}$ & $\mathtt{GAMAN}$  \tabularnewline
 $L_2=0$ &  & $L_2=7957$  &  $L_2=1000$  &  $L_2=52$  &  $L_2=1000$  & $L_2=2000$  & $L_2=3000$  \tabularnewline
\includegraphics[width=0.25\columnwidth]{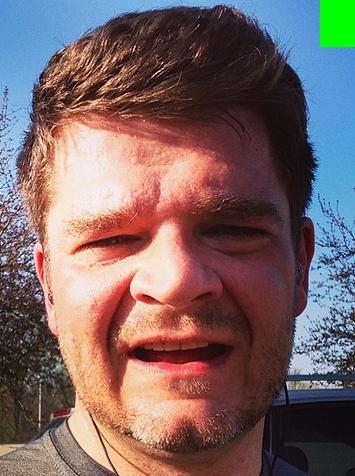} & &
\includegraphics[width=0.25\columnwidth]{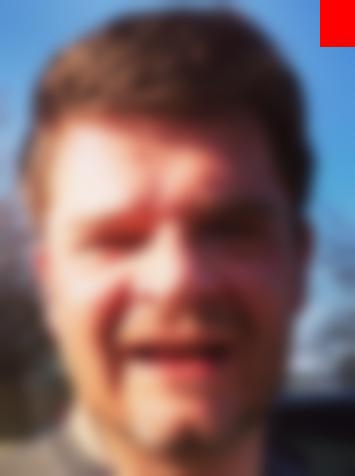} & 
\includegraphics[width=0.25\columnwidth]{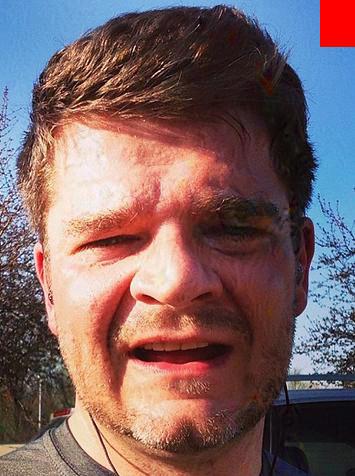} & 
\includegraphics[width=0.25\columnwidth]{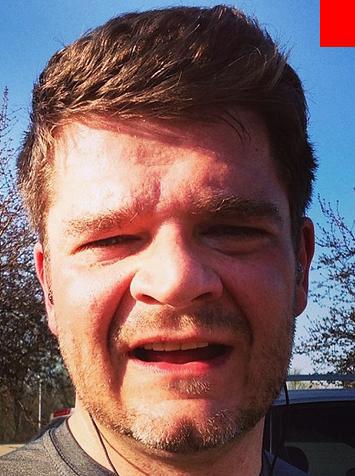} & 
\includegraphics[width=0.25\columnwidth]{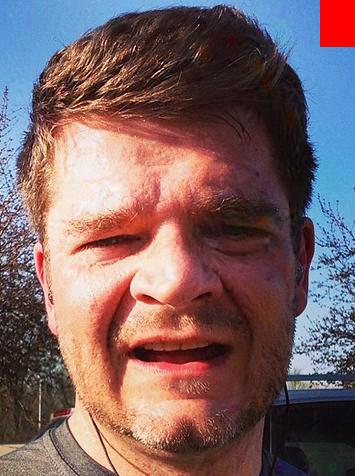} & 
\includegraphics[width=0.25\columnwidth]{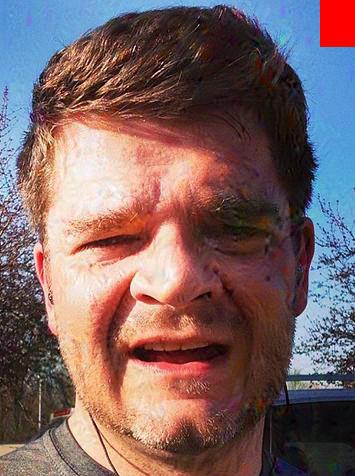} & 
\includegraphics[width=0.25\columnwidth]{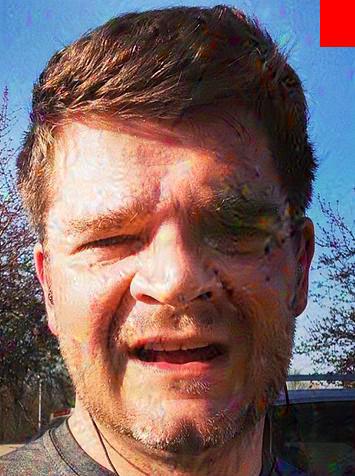} \tabularnewline
 & & & 
\includegraphics[width=0.25\columnwidth]{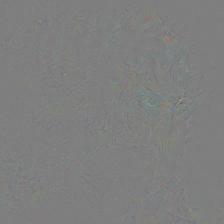} & 
\includegraphics[width=0.25\columnwidth]{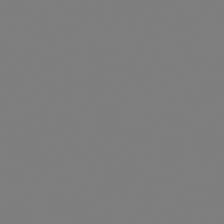} & 
\includegraphics[width=0.25\columnwidth]{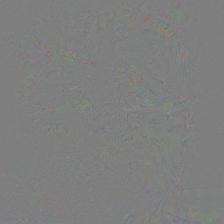} & 
\includegraphics[width=0.25\columnwidth]{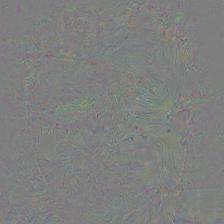} & 
\includegraphics[width=0.25\columnwidth]{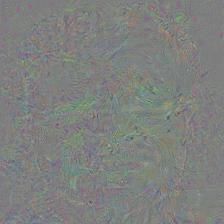} \tabularnewline
\\
 $L_2=0$ &  & $L_2=5071$  &  $L_2=1000$  &  $L_2=185$  &  $L_2=1000$  & $L_2=2000$  & $L_2=3000$  \tabularnewline
\includegraphics[width=0.25\columnwidth]{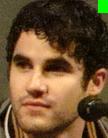} & &
\includegraphics[width=0.25\columnwidth]{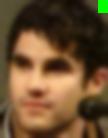} & 
\includegraphics[width=0.25\columnwidth]{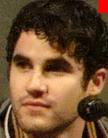} & 
\includegraphics[width=0.25\columnwidth]{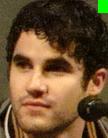} & 
\includegraphics[width=0.25\columnwidth]{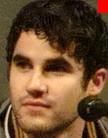} & 
\includegraphics[width=0.25\columnwidth]{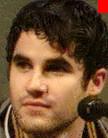} & 
\includegraphics[width=0.25\columnwidth]{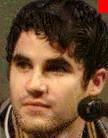} \tabularnewline
 & & & 
\includegraphics[width=0.25\columnwidth]{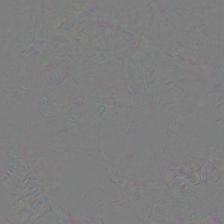} & 
\includegraphics[width=0.25\columnwidth]{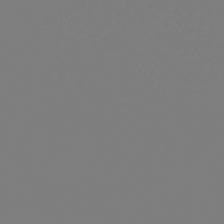} & 
\includegraphics[width=0.25\columnwidth]{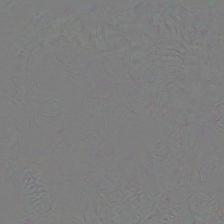} & 
\includegraphics[width=0.25\columnwidth]{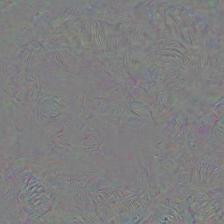} & 
\includegraphics[width=0.25\columnwidth]{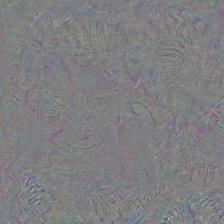} \tabularnewline
\\
 $L_2=0$ &  & $L_2=5123$  &  $L_2=1000$  &  $L_2=144$  &  $L_2=1000$  & $L_2=2000$  & $L_2=3000$  \tabularnewline
\includegraphics[width=0.25\columnwidth]{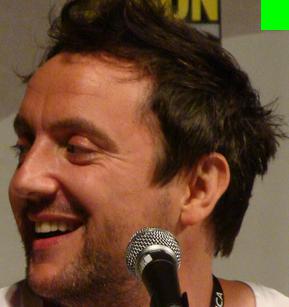} & &
\includegraphics[width=0.25\columnwidth]{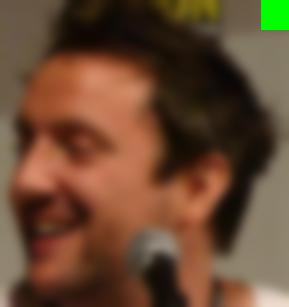} & 
\includegraphics[width=0.25\columnwidth]{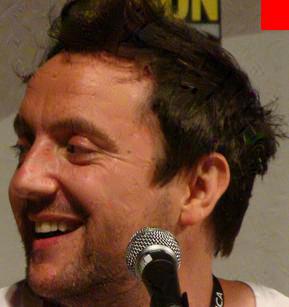} & 
\includegraphics[width=0.25\columnwidth]{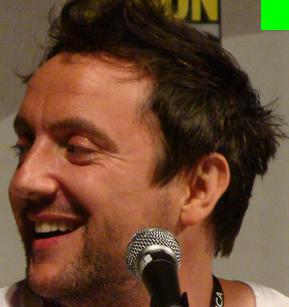} & 
\includegraphics[width=0.25\columnwidth]{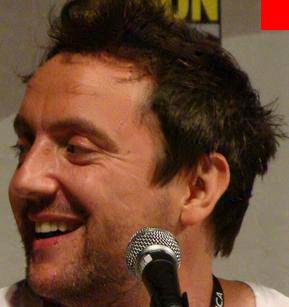} & 
\includegraphics[width=0.25\columnwidth]{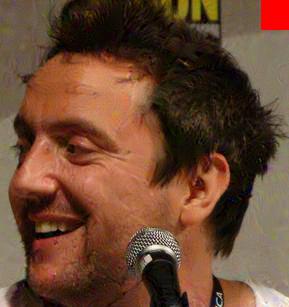} & 
\includegraphics[width=0.25\columnwidth]{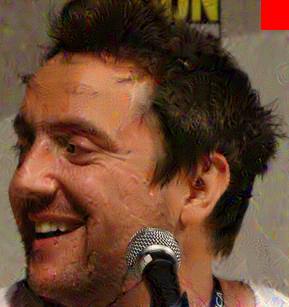} \tabularnewline
 & & & 
\includegraphics[width=0.25\columnwidth]{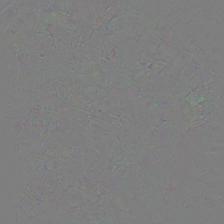} & 
\includegraphics[width=0.25\columnwidth]{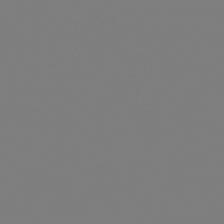} & 
\includegraphics[width=0.25\columnwidth]{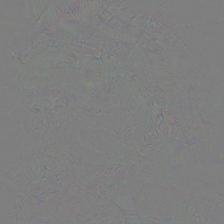} & 
\includegraphics[width=0.25\columnwidth]{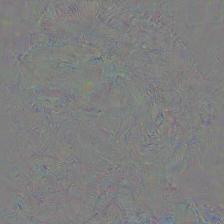} & 
\includegraphics[width=0.25\columnwidth]{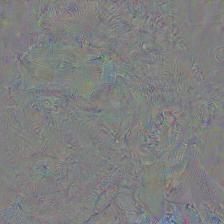} \tabularnewline

\end{tabular}
\par\end{centering}{\footnotesize \par}
\vspace{0em}

\caption{\label{fig:example2}Randomly chosen perturbed images after $\mathsf{Proc}$ and the corresponding GoogleNet predictions (green for correct, red for wrong). Perturbations are visualised with gray background. $\mathtt{GA}$ and $\mathtt{GAMAN}$ reliably confuse the classifier at almost no cost on the aesthetics. As the $L_2$ norm increases, artifacts become more visible. Perturbations may be too small to be visible when printed; zoom in in electronic version for better visibility.}
\end{figure*}

\begin{figure*}[t]
\begin{centering}
{\footnotesize{}}%
\begin{tabular}{cccccccc}
Original & & Blur & $\mathtt{GA}$ & $\mathtt{DF}\cite{moosavi2016cvpr}$ & $\mathtt{GAMAN}$ & $\mathtt{GAMAN}$ & $\mathtt{GAMAN}$  \tabularnewline
 $L_2=0$ &  & $L_2=5365$  &  $L_2=1000$  &  $L_2=138$  &  $L_2=1000$  & $L_2=2000$  & $L_2=3000$  \tabularnewline
\includegraphics[width=0.25\columnwidth]{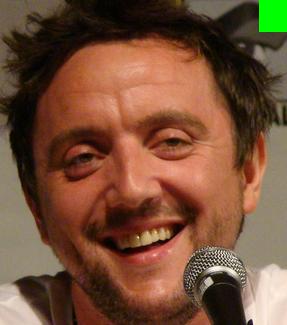} & &
\includegraphics[width=0.25\columnwidth]{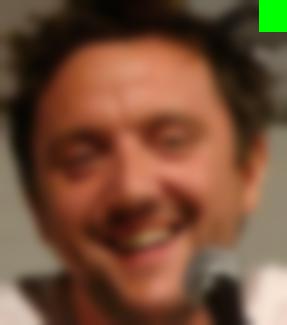} & 
\includegraphics[width=0.25\columnwidth]{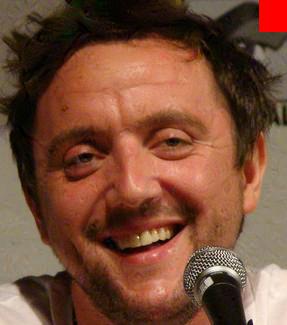} & 
\includegraphics[width=0.25\columnwidth]{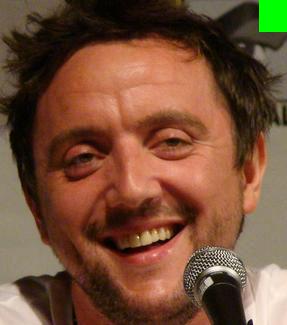} & 
\includegraphics[width=0.25\columnwidth]{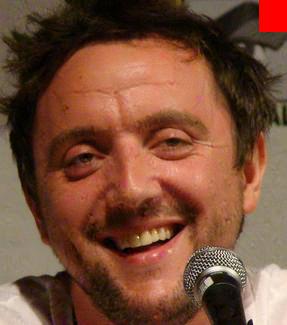} & 
\includegraphics[width=0.25\columnwidth]{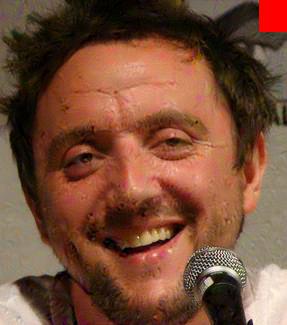} & 
\includegraphics[width=0.25\columnwidth]{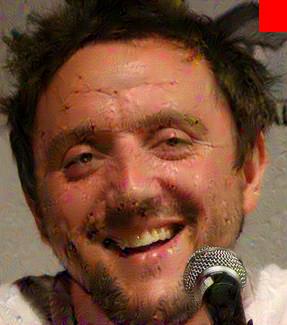} \tabularnewline
 & & & 
\includegraphics[width=0.25\columnwidth]{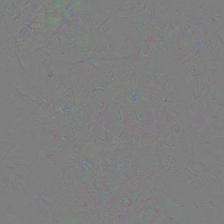} & 
\includegraphics[width=0.25\columnwidth]{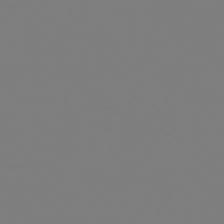} & 
\includegraphics[width=0.25\columnwidth]{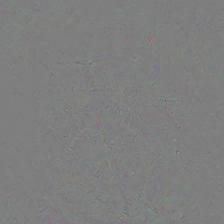} & 
\includegraphics[width=0.25\columnwidth]{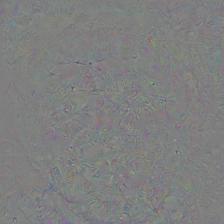} & 
\includegraphics[width=0.25\columnwidth]{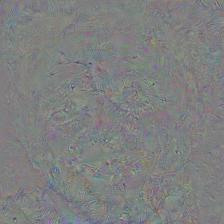} \tabularnewline
\\
 $L_2=0$ &  & $L_2=3813$  &  $L_2=1000$  &  $L_2=89$  &  $L_2=1000$  & $L_2=2000$  & $L_2=3000$  \tabularnewline
\includegraphics[width=0.25\columnwidth]{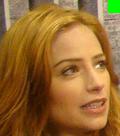} & &
\includegraphics[width=0.25\columnwidth]{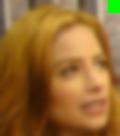} & 
\includegraphics[width=0.25\columnwidth]{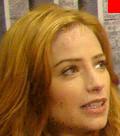} & 
\includegraphics[width=0.25\columnwidth]{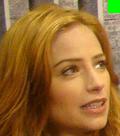} & 
\includegraphics[width=0.25\columnwidth]{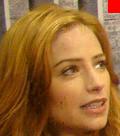} & 
\includegraphics[width=0.25\columnwidth]{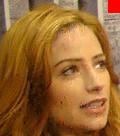} & 
\includegraphics[width=0.25\columnwidth]{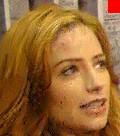} \tabularnewline
 & & & 
\includegraphics[width=0.25\columnwidth]{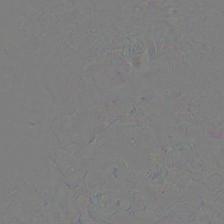} & 
\includegraphics[width=0.25\columnwidth]{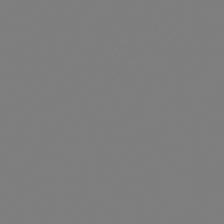} & 
\includegraphics[width=0.25\columnwidth]{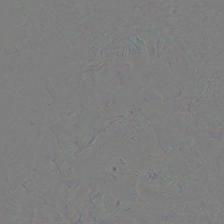} & 
\includegraphics[width=0.25\columnwidth]{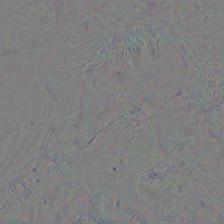} & 
\includegraphics[width=0.25\columnwidth]{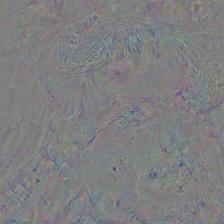} \tabularnewline
\\
 $L_2=0$ &  & $L_2=6539$  &  $L_2=1000$  &  $L_2=113$  &  $L_2=1000$  & $L_2=2000$  & $L_2=3000$  \tabularnewline
\includegraphics[width=0.25\columnwidth]{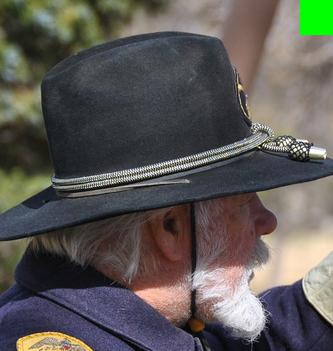} & &
\includegraphics[width=0.25\columnwidth]{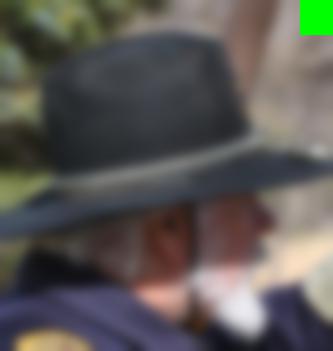} & 
\includegraphics[width=0.25\columnwidth]{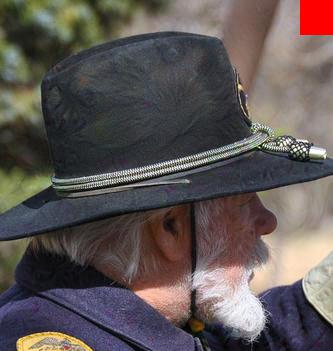} & 
\includegraphics[width=0.25\columnwidth]{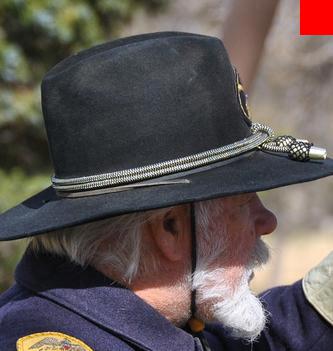} & 
\includegraphics[width=0.25\columnwidth]{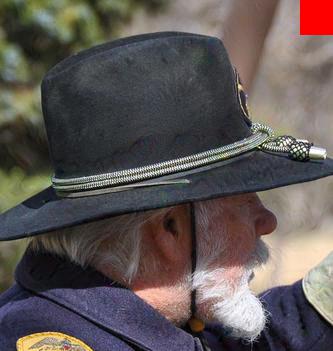} & 
\includegraphics[width=0.25\columnwidth]{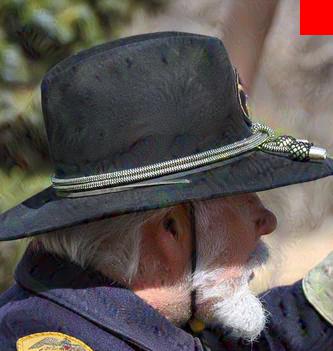} & 
\includegraphics[width=0.25\columnwidth]{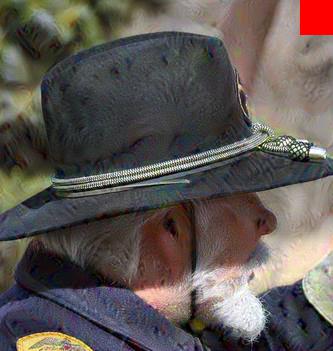} \tabularnewline
 & & & 
\includegraphics[width=0.25\columnwidth]{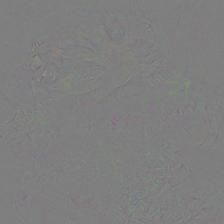} & 
\includegraphics[width=0.25\columnwidth]{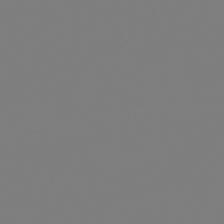} & 
\includegraphics[width=0.25\columnwidth]{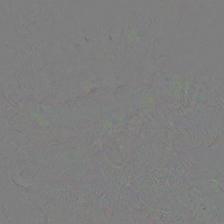} & 
\includegraphics[width=0.25\columnwidth]{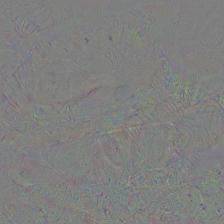} & 
\includegraphics[width=0.25\columnwidth]{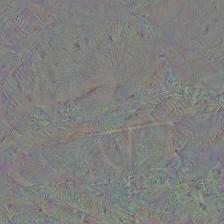} \tabularnewline

\end{tabular}
\par\end{centering}{\footnotesize \par}
\vspace{0em}

\caption{\label{fig:example3}Randomly chosen perturbed images after $\mathsf{Proc}$ and the corresponding GoogleNet predictions (green for correct, red for wrong). Perturbations are visualised with gray background. $\mathtt{GA}$ and $\mathtt{GAMAN}$ reliably confuse the classifier at almost no cost on the aesthetics. As the $L_2$ norm increases, artifacts become more visible. Perturbations may be too small to be visible when printed; zoom in in electronic version for better visibility.}
\end{figure*}

\begin{figure*}[t]
\begin{centering}
{\footnotesize{}}%
\begin{tabular}{cccccccc}
Original & & Blur & $\mathtt{GA}$ & $\mathtt{DF}\cite{moosavi2016cvpr}$ & $\mathtt{GAMAN}$ & $\mathtt{GAMAN}$ & $\mathtt{GAMAN}$  \tabularnewline
 $L_2=0$ &  & $L_2=2586$  &  $L_2=1000$  &  $L_2=75$  &  $L_2=1000$  & $L_2=2000$  & $L_2=3000$  \tabularnewline
\includegraphics[width=0.25\columnwidth]{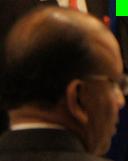} & &
\includegraphics[width=0.25\columnwidth]{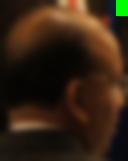} & 
\includegraphics[width=0.25\columnwidth]{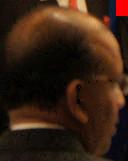} & 
\includegraphics[width=0.25\columnwidth]{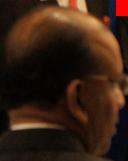} & 
\includegraphics[width=0.25\columnwidth]{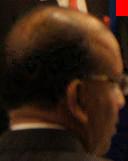} & 
\includegraphics[width=0.25\columnwidth]{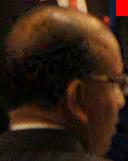} & 
\includegraphics[width=0.25\columnwidth]{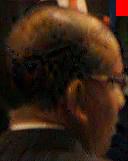} \tabularnewline
 & & & 
\includegraphics[width=0.25\columnwidth]{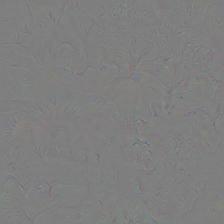} & 
\includegraphics[width=0.25\columnwidth]{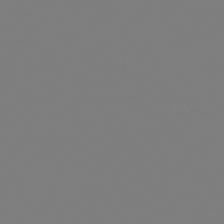} & 
\includegraphics[width=0.25\columnwidth]{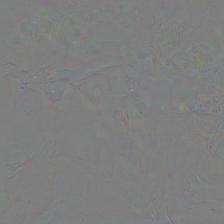} & 
\includegraphics[width=0.25\columnwidth]{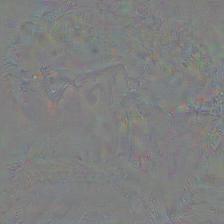} & 
\includegraphics[width=0.25\columnwidth]{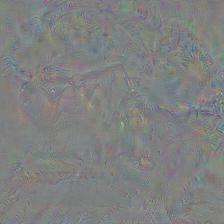} \tabularnewline
\\
 $L_2=0$ &  & $L_2=4855$  &  $L_2=1000$  &  $L_2=153$  &  $L_2=1000$  & $L_2=2000$  & $L_2=3000$  \tabularnewline
\includegraphics[width=0.25\columnwidth]{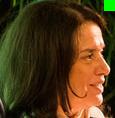} & &
\includegraphics[width=0.25\columnwidth]{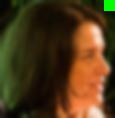} & 
\includegraphics[width=0.25\columnwidth]{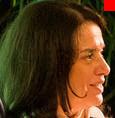} & 
\includegraphics[width=0.25\columnwidth]{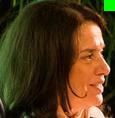} & 
\includegraphics[width=0.25\columnwidth]{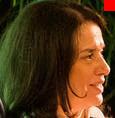} & 
\includegraphics[width=0.25\columnwidth]{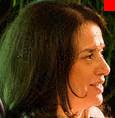} & 
\includegraphics[width=0.25\columnwidth]{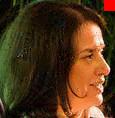} \tabularnewline
 & & & 
\includegraphics[width=0.25\columnwidth]{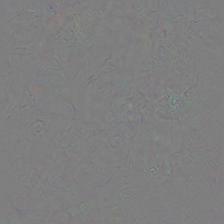} & 
\includegraphics[width=0.25\columnwidth]{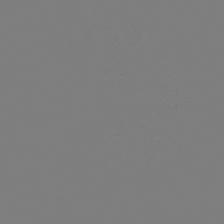} & 
\includegraphics[width=0.25\columnwidth]{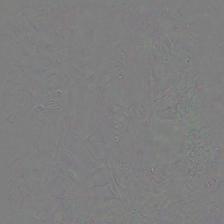} & 
\includegraphics[width=0.25\columnwidth]{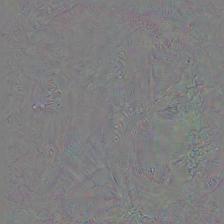} & 
\includegraphics[width=0.25\columnwidth]{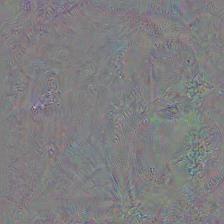} \tabularnewline
\\
 $L_2=0$ &  & $L_2=5018$  &  $L_2=1000$  &  $L_2=0$  &  $L_2=1000$  & $L_2=2000$  & $L_2=3000$  \tabularnewline
\includegraphics[width=0.25\columnwidth]{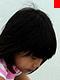} & &
\includegraphics[width=0.25\columnwidth]{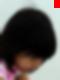} & 
\includegraphics[width=0.25\columnwidth]{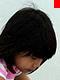} & 
\includegraphics[width=0.25\columnwidth]{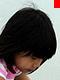} & 
\includegraphics[width=0.25\columnwidth]{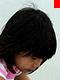} & 
\includegraphics[width=0.25\columnwidth]{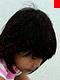} & 
\includegraphics[width=0.25\columnwidth]{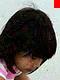} \tabularnewline
 & & & 
\includegraphics[width=0.25\columnwidth]{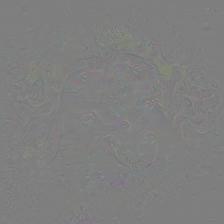} & 
\includegraphics[width=0.25\columnwidth]{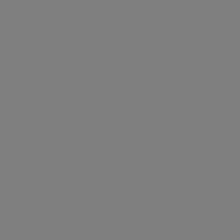} & 
\includegraphics[width=0.25\columnwidth]{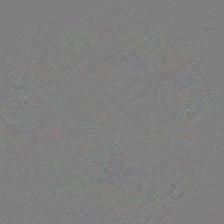} & 
\includegraphics[width=0.25\columnwidth]{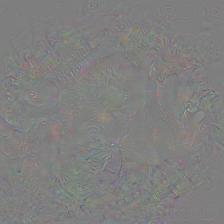} & 
\includegraphics[width=0.25\columnwidth]{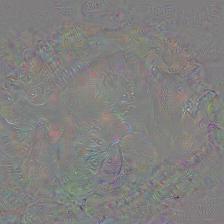} \tabularnewline

\end{tabular}
\par\end{centering}{\footnotesize \par}
\vspace{0em}

\caption{\label{fig:example4}Randomly chosen perturbed images after $\mathsf{Proc}$ and the corresponding GoogleNet predictions (green for correct, red for wrong). Perturbations are visualised with gray background. $\mathtt{GA}$ and $\mathtt{GAMAN}$ reliably confuse the classifier at almost no cost on the aesthetics. As the $L_2$ norm increases, artifacts become more visible. Perturbations may be too small to be visible when printed; zoom in in electronic version for better visibility.}
\end{figure*}

\noindent
We include more qualitative results (equivalent to figure 3 in the main paper). See figures \ref{fig:example1}, \ref{fig:example2}, \ref{fig:example3}, \ref{fig:example4}.

\begin{table}[H]
\begin{centering}
\setlength\tabcolsep{0.3em}
\begin{tabular}{cccc}
 
\multirow{2}{*}{Network} &  & \multirow{2}{*}{Optimal Strategy $\theta^{u\star}$} & Bound on \tabularnewline
& & & Rec. Rate \tabularnewline

\cline{1-1} \cline{3-4} 
\vspace{-1em}  &  &  &  \tabularnewline
\cline{1-1} \cline{3-4} 
\vspace{-1em}  &  &  &  \tabularnewline
AlexNet & & $(/\mathtt{B}:100\%)$ & $\leq$6.4 \tabularnewline
\vspace{-1em}  &  &  & \tabularnewline
VGG & & $(/\mathtt{B}:86\%,/\mathtt{TNBC}:14\%)$ & $\leq$4.9 \tabularnewline
\vspace{-1em} &  &  & \tabularnewline
GoogleNet & & $(/\mathtt{B}:61\%,/\mathtt{TNBC}:39\%)$ & $\leq$7.3 \tabularnewline
\vspace{-1em}  &  &  & \tabularnewline
ResNet & & $(/\mathtt{B}:31\%,/\mathtt{TNBC}:69\%)$ & $\leq$8.5 \tabularnewline
\vspace{-1em}
 &  &  &  \tabularnewline
\cline{1-1} \cline{3-4}
\end{tabular}
\par\end{centering}
\caption{\label{tab:supp-opt}Optimal strategies and the corresponding guaranteed upper bounds on the recognition rate for different networks. We write $\leq\cdot\,\,$ to denote the upper bound.}
\end{table}

\newpage

\begin{table}
\newbox\minimaxunderlined
\sbox\minimaxunderlined{\ul{13.5}}
\begin{centering}
\setlength\tabcolsep{0.4em}
\begin{tabular}{ccccccccc}
 & & \multicolumn{7}{c}{AlexNet} \\
\vspace{-1em} &  &  &  &  &  &  &  & \tabularnewline
Perturb &  & $\emptyset$ & $\mathsf{Proc}$ & $\mathsf{T}$ & $\mathsf{N}$ & $\mathsf{B}$ & $\mathsf{C}$ & {\footnotesize $\mathsf{TNBC}$}\tabularnewline
\cline{1-1} \cline{3-9} 
\vspace{-1em}
 &  &  &  &  &  &  &  & \tabularnewline
\cline{1-1} \cline{3-9} 
\vspace{-.7em}
 &  &  &  &  &  &  &  & \tabularnewline
None &  & \hl{83.8} & \hl{83.8} & 83.7 & 77.8 & 78.7 & 80.1 & 83.9 \tabularnewline
\cline{1-1} \cline{3-9} 
\vspace{-1em}
 &  &  &  &  &  &  &  & \tabularnewline
$\mathtt{BI}$\cite{kurakin2016adversarial} &  & 1.2 & 10.0 & 29.7 & 20.8 & 26.6 & \hl{34.3} & 23.3 \tabularnewline
$\mathtt{GA}$ &  & 0.2 & 4.8 & 13.6 & 11.6 & 17.7 & \hl{17.8} & 12.2 \tabularnewline
$\mathtt{BI}\text{-}\mathtt{S}$\cellcolor{Gray} &  & 1.2 & 10.1 & 31.2 & 21.0 & 27.2 & \hl{35.7} & 23.3 \tabularnewline
$\mathtt{GA}\text{-}\mathtt{S}$\cellcolor{Gray} &  & 0.2 & 5.0 & 15.4 & 12.6 & 19.0 & \hl{19.3} & 12.8 \tabularnewline
\vspace{-1.2em}
 &  &  &  &  &  &  &  & \tabularnewline
\cline{1-1} \cline{3-9} 
\vspace{-1em}
 &  &  &  &  &  &  &  & \tabularnewline
$\mathtt{DF}$\cite{moosavi2016cvpr} &  & \ul{0.0} & 62.1 & 76.5 & 68.5 & 69.4 & \hl{75.0} & 74.7 \tabularnewline
$\mathtt{GAMAN}$ &  & \ul{0.0} & \ul{1.4} & \ul{6.4} & \ul{9.2} &  \hl{\mbox{\usebox\minimaxunderlined}} & \ul{12.3} & \ul{5.6} \tabularnewline
\cline{1-1} \cline{3-9} 
\end{tabular}
\par\end{centering}
\vspace{1em}

\newbox\minimaxunderlined
\sbox\minimaxunderlined{\ul{11.8}}
\begin{centering}
\setlength\tabcolsep{0.4em}
\begin{tabular}{ccccccccc}
 & & \multicolumn{7}{c}{VGG} \\
\vspace{-1em} &  &  &  &  &  &  &  & \tabularnewline
Perturb &  & $\emptyset$ & $\mathsf{Proc}$ & $\mathsf{T}$ & $\mathsf{N}$ & $\mathsf{B}$ & $\mathsf{C}$ & {\footnotesize $\mathsf{TNBC}$}\tabularnewline
\cline{1-1} \cline{3-9} 
\vspace{-1em}
 &  &  &  &  &  &  &  & \tabularnewline
\cline{1-1} \cline{3-9} 
\vspace{-.7em}
 &  &  &  &  &  &  &  & \tabularnewline
None &  & \hl{86.1} & \hl{86.1} & 84.8 & 77.2 & 81.5 & 84.1 & 85.8 \tabularnewline
\cline{1-1} \cline{3-9} 
\vspace{-1em}
 &  &  &  &  &  &  &  & \tabularnewline
$\mathtt{BI}$\cite{kurakin2016adversarial} &  & 0.5 & 6.8 & 11.1 & 18.1 & \hl{23.2} & 16.8 & 14.4 \tabularnewline
$\mathtt{GA}$ &  & \ul{0.0} & 4.2 & 5.5 & 11.2 & \hl{17.2} & 10.2 & 8.2 \tabularnewline
$\mathtt{BI}\text{-}\mathtt{S}$\cellcolor{Gray} &  & 0.3 & 7.1 & 11.2 & 19.2 & \hl{23.8} & 17.3 & 14.3 \tabularnewline
$\mathtt{GA}\text{-}\mathtt{S}$\cellcolor{Gray} &  & \ul{0.0} & 4.8 & 5.9 & 11.9 & \hl{18.6} & 11.3 & 8.8 \tabularnewline
\vspace{-1.2em}
 &  &  &  &  &  &  &  & \tabularnewline
\cline{1-1} \cline{3-9} 
\vspace{-1em}
 &  &  &  &  &  &  &  & \tabularnewline
$\mathtt{DF}$\cite{moosavi2016cvpr} &  & \ul{0.0} & 53.3 & 66.3 & 65.9 & 69.4 & 69.2 & \hl{71.4} \tabularnewline
$\mathtt{GAMAN}$ &  & \ul{0.0} & \ul{1.6} & \ul{2.1} & \ul{8.5} &  \hl{\mbox{\usebox\minimaxunderlined}} & \ul{5.6} & \ul{3.5} \tabularnewline
\cline{1-1} \cline{3-9} 
\end{tabular}
\par\end{centering}
\vspace{1em}

\newbox\minimaxunderlined
\sbox\minimaxunderlined{\ul{22.2}}
\begin{centering}
\setlength\tabcolsep{0.4em}
\begin{tabular}{ccccccccc}
 & & \multicolumn{7}{c}{GoogleNet} \\
Perturb &  & $\emptyset$ & $\mathsf{Proc}$ & $\mathsf{T}$ & $\mathsf{N}$ & $\mathsf{B}$ & $\mathsf{C}$ & {\footnotesize $\mathsf{TNBC}$}\tabularnewline
\cline{1-1} \cline{3-9} 
\vspace{-1em}
 &  &  &  &  &  &  &  & \tabularnewline
\cline{1-1} \cline{3-9} 
\vspace{-.7em}
 &  &  &  &  &  &  &  & \tabularnewline
None &  & \hl{87.8} & \hl{87.8} & 87.6 & 64.0 & 81.2 & 85.4 & 87.3\tabularnewline
\cline{1-1} \cline{3-9} 
\vspace{-1em}
 &  &  &  &  &  &  &  & \tabularnewline
$\mathtt{BI}$\cite{kurakin2016adversarial} &  & \ul{0.0} & 8.3 & 15.8 & 16.8 & \hl{28.6} & 27.4 & 17.6 \tabularnewline
$\mathtt{GA}$ &  & \ul{0.0} & 8.6 & 13.2 & \ul{14.1} & \hl{28.4} & 23.7 & 16.4 \tabularnewline
$\mathtt{BI}\text{-}\mathtt{S}$\cellcolor{Gray} &  & \ul{0.0} & 8.8 & 17.2 & 17.7 & \hl{29.3} & 28.8 & 18.8 \tabularnewline
$\mathtt{GA}\text{-}\mathtt{S}$\cellcolor{Gray} &  & \ul{0.0} & 9.1 & 14.9 & 15.2 & \hl{29.3} & 25.5 & 18.0 \tabularnewline
\vspace{-1.2em}
 &  &  &  &  &  &  &  & \tabularnewline
\cline{1-1} \cline{3-9} 
\vspace{-1em}
 &  &  &  &  &  &  &  & \tabularnewline
$\mathtt{DF}$\cite{moosavi2016cvpr} &  & \ul{0.0} & 51.8 & 75.6 & 56.5 & 72.5 & \hl{76.9} & 75.5\tabularnewline
$\mathtt{GAMAN}$ &  & \ul{0.0} & \ul{4.0} & \ul{6.6} & 15.0 & \hl{\mbox{\usebox\minimaxunderlined}} & \ul{16.7} & \ul{9.9} \tabularnewline
\cline{1-1} \cline{3-9} 
\end{tabular}
\par\end{centering}
\vspace{1em}

\newbox\minimaxunderlined
\sbox\minimaxunderlined{\ul{31.8}}
\begin{centering}
\setlength\tabcolsep{0.4em}
\begin{tabular}{ccccccccc}
 & & \multicolumn{7}{c}{ResNet} \\
\vspace{-1em} &  &  &  &  &  &  &  & \tabularnewline
Perturb &  & $\emptyset$ & $\mathsf{Proc}$ & $\mathsf{T}$ & $\mathsf{N}$ & $\mathsf{B}$ & $\mathsf{C}$ & {\footnotesize $\mathsf{TNBC}$}\tabularnewline
\cline{1-1} \cline{3-9} 
\vspace{-1em}
 &  &  &  &  &  &  &  & \tabularnewline
\cline{1-1} \cline{3-9} 
\vspace{-.7em}
 &  &  &  &  &  &  &  & \tabularnewline
None &  & \hl{91.1} & \hl{91.1} & 90.6 & 72.0 & 87.2 & 89.3 & 90.8\tabularnewline
\cline{1-1} \cline{3-9} 
\vspace{-1em}
 &  &  &  &  &  &  &  & \tabularnewline
$\mathtt{BI}$\cite{kurakin2016adversarial} &  & \ul{0.0} & 10.9 & 36.8 & 24.8 & 32.8 & \hl{45.3} & 26.3 \tabularnewline
$\mathtt{GA}$ &  & \ul{0.0} & 15.2 & 37.3 & 24.4 & 36.9 & \hl{43.7} & 28.9 \tabularnewline
$\mathtt{BI}\text{-}\mathtt{S}$\cellcolor{Gray} &  & \ul{0.0} & 13.0 & 43.4 & 27.4 & 35.8 & \hl{51.5} & 29.9 \tabularnewline
$\mathtt{GA}\text{-}\mathtt{S}$\cellcolor{Gray} &  & \ul{0.0} & 19.4 & 45.0 & 27.1 & 40.2 & \hl{50.3} & 33.3 \tabularnewline
\vspace{-1.2em}
 &  &  &  &  &  &  &  & \tabularnewline
\cline{1-1} \cline{3-9} 
\vspace{-1em}
 &  &  &  &  &  &  &  & \tabularnewline
$\mathtt{DF}$\cite{moosavi2016cvpr} &  & \ul{0.0} & 52.9 & 83.1 & 65.0 & 76.8 & \hl{84.2} & 80.9 \tabularnewline
$\mathtt{GAMAN}$ &  & \ul{0.0} & \ul{7.3} & \ul{23.4} & \ul{23.3} & \ul{28.2} &  \hl{\mbox{\usebox\minimaxunderlined}} & \ul{18.4} \tabularnewline
\cline{1-1} \cline{3-9} 
\end{tabular}
\par\end{centering}
\vspace{1em}

\caption{\label{tab:supp-multiarch-robustness}Extended version of table \ref{tab:adv-robustness} in the main paper for all four network architectures; additional AIP entries are denoted as gray cells. Robustness analysis of AIPs for various convnet architectures. AIPs are restricted to $||\cdot ||_2 \leq 1000$. $(\mathsf{T},\mathsf{N},\mathsf{B},\mathsf{C})=$ (Translate, Noise, Blur, Crop).}
\end{table}

\begin{table}




\newbox\minimaxunderlined
\sbox\minimaxunderlined{\ul{6.4}}
\begin{centering}
\setlength\tabcolsep{0.5em}
\begin{tabular}{cccccccc}
& &  \multicolumn{6}{c}{AlexNet}\tabularnewline
\vspace{-1em}
 &  &  &  &  &  &  & \tabularnewline
 &  & \multicolumn{6}{c}{Recogniser $\Theta^r$}\tabularnewline
\vspace{-1em}
 &  &  &  &  &  &  & \tabularnewline
\cline{3-8} 
\vspace{-1em}
 &  &  &  &  &  &  & \tabularnewline
User $\Theta^u$ &  & $\mathsf{Proc}$ & $\mathsf{T}$ & $\mathsf{N}$ & $\mathsf{B}$ & $\mathsf{C}$ & {\footnotesize $\mathsf{TNBC}$}\tabularnewline
\vspace{-1em}
 &  &  &  &  &  &  & \tabularnewline
\cline{1-1} \cline{3-8} 
\vspace{-1em}
 &  &  &  &  &  &  & \tabularnewline
\cline{1-1} \cline{3-8} 
\vspace{-1em}
 &  &  &  &  &  &  & \tabularnewline
$\mathtt{GAMAN}$ &  & 1.4 & 6.4 & 9.2 & \hl{13.5} & 12.3 & 5.6 \tabularnewline
$\mathtt{/T}$ &  & 0.9 & \ul{0.8} & 6.2 & \hl{10.5} & 2.7 & 2.2  \tabularnewline
$\mathtt{/N}$ &  & 1.2 & 4.2 & 4.8 & \hl{11.7} & 9.5 & 3.9  \tabularnewline
$\mathtt{/B}$ &  & 0.8 & 3.5 & 6.3 & \hl{\mbox{\usebox\minimaxunderlined}} & 6.0 & 2.6 \tabularnewline
$\mathtt{/C}$ &  & 2.4 & 2.5 & 9.2 & \hl{13.1} & \ul{1.3} & 3.4 \tabularnewline
$\mathtt{/TNBC}$ &  & \ul{0.6} & 1.2 & \ul{4.5} & \hl{7.8} & 2.9 & \ul{1.9} \tabularnewline
\vspace{-1em}
 &  &  &  &  &  &  &\tabularnewline
\cline{1-1} \cline{3-8} 
\end{tabular}
\par\end{centering}
\vspace{1em}




\begin{centering}
\setlength\tabcolsep{0.5em}
\begin{tabular}{cccccccc}
& &  \multicolumn{6}{c}{VGG}\tabularnewline
\vspace{-1em}
 &  &  &  &  &  &  & \tabularnewline
 &  & \multicolumn{6}{c}{Recogniser $\Theta^r$}\tabularnewline
\vspace{-1em}
 &  &  &  &  &  &  & \tabularnewline
\cline{3-8} 
\vspace{-1em}
 &  &  &  &  &  &  & \tabularnewline
User $\Theta^u$ &  & $\mathsf{Proc}$ & $\mathsf{T}$ & $\mathsf{N}$ & $\mathsf{B}$ & $\mathsf{C}$ & {\footnotesize $\mathsf{TNBC}$}\tabularnewline
\vspace{-1em}
 &  &  &  &  &  &  & \tabularnewline
\cline{1-1} \cline{3-8} 
\vspace{-1em}
 &  &  &  &  &  &  & \tabularnewline
\cline{1-1} \cline{3-8} 
\vspace{-1em}
 &  &  &  &  &  &  & \tabularnewline
$\mathtt{GAMAN}$ &  & 1.6 & 2.1 & 8.5 & \hl{11.8} & 5.6 & 3.5 \tabularnewline
$\mathtt{/T}$ &  & 1.5 & 1.2 & 8.1 & \hl{12.3} & 3.2 & 2.8    \tabularnewline
$\mathtt{/N}$ &  & 2.0 & 2.5 & \ul{3.9} & \hl{12.6} & 6.7 & 3.9  \tabularnewline
$\mathtt{/B}$ &  & \ul{0.3} & \ul{0.7} & \hl{5.0} & \ul{4.5} & 2.2 & \ul{1.2}  \tabularnewline
$\mathtt{/C}$ &  & 2.0 & 1.6 & 9.5 & \hl{14.0} & \ul{1.9} & 3.1 \tabularnewline
$\mathtt{/TNBC}$ &  & 0.6 & \ul{0.7} & 4.3 & \hl{7.3} & 2.3 & 1.4\tabularnewline
\vspace{-1em}
 &  &  &  &  &  &  &\tabularnewline
\cline{1-1} \cline{3-8} 
\end{tabular}
\par\end{centering}
\vspace{1em}




\begin{centering}
\setlength\tabcolsep{0.5em}
\begin{tabular}{cccccccc}
& &  \multicolumn{6}{c}{GoogleNet}\tabularnewline
\vspace{-1em}
 &  &  &  &  &  &  & \tabularnewline
 &  & \multicolumn{6}{c}{Recogniser $\Theta^r$}\tabularnewline
\vspace{-1em}
 &  &  &  &  &  &  & \tabularnewline
\cline{3-8} 
\vspace{-1em}
 &  &  &  &  &  &  & \tabularnewline
User $\Theta^u$ &  & $\mathsf{Proc}$ & $\mathsf{T}$ & $\mathsf{N}$ & $\mathsf{B}$ & $\mathsf{C}$ & {\footnotesize $\mathsf{TNBC}$}\tabularnewline
\vspace{-1em}
 &  &  &  &  &  &  & \tabularnewline
\cline{1-1} \cline{3-8} 
\vspace{-1em}
 &  &  &  &  &  &  & \tabularnewline
\cline{1-1} \cline{3-8} 
\vspace{-1em}
 &  &  &  &  &  &  & \tabularnewline
$\mathtt{GAMAN}$ &  & 4.0 & 6.6 & 15.0 & \hl{22.2} & 16.7 & 9.9 \tabularnewline
$\mathtt{/T}$ &  & 2.5 & 2.3 & 11.6 & \hl{18.5} & 7.2 & 4.9 \tabularnewline
$\mathtt{/N}$ &  & 5.8 & 7.6 & \ul{4.6} & \hl{23.6} & 16.6 & 9.1 \tabularnewline
$\mathtt{/B}$ &  & \ul{0.4} & \ul{0.8} & \hl{8.6} & \ul{5.8} & \ul{3.1} & \ul{1.4} \tabularnewline
$\mathtt{/C}$ &  & 2.6 & 2.2 & 11.8 & \hl{18.1} & 3.4 & 4.3 \tabularnewline
$\mathtt{/TNBC}$ &  & 0.7 & 0.9 & 5.2 & \hl{9.5} & 3.2 & 2.0 \tabularnewline
\vspace{-1em}
 &  &  &  &  &  &  &\tabularnewline
\cline{1-1} \cline{3-8} 
\end{tabular}
\par\end{centering}
\vspace{1em}




\begin{centering}
\setlength\tabcolsep{0.5em}
\begin{tabular}{cccccccc}
& &  \multicolumn{6}{c}{ResNet}\tabularnewline
\vspace{-1em}
 &  &  &  &  &  &  & \tabularnewline
 &  & \multicolumn{6}{c}{Recogniser $\Theta^r$}\tabularnewline
\vspace{-1em}
 &  &  &  &  &  &  & \tabularnewline
\cline{3-8} 
\vspace{-1em}
 &  &  &  &  &  &  & \tabularnewline
User $\Theta^u$ &  & $\mathsf{Proc}$ & $\mathsf{T}$ & $\mathsf{N}$ & $\mathsf{B}$ & $\mathsf{C}$ & {\footnotesize $\mathsf{TNBC}$}\tabularnewline
\vspace{-1em}
 &  &  &  &  &  &  & \tabularnewline
\cline{1-1} \cline{3-8} 
\vspace{-1em}
 &  &  &  &  &  &  & \tabularnewline
\cline{1-1} \cline{3-8} 
\vspace{-1em}
 &  &  &  &  &  &  & \tabularnewline
$\mathtt{GAMAN}$ &  & 7.3 & 23.4 & 23.3 & 28.2 & \hl{31.8} & 18.4 \tabularnewline
$\mathtt{/T}$ &  & 2.9 & 2.8 & 16.6 & \hl{19.0} & 5.4 & 5.8  \tabularnewline
$\mathtt{/N}$ &  & 5.3 & 12.9 & \ul{4.2} & \hl{23.5} & 20.1 & 10.2 \tabularnewline
$\mathtt{/B}$ &  & \ul{0.6} & 3.1 & \hl{13.0} & \ul{6.8} & 5.3 & 2.4\tabularnewline
$\mathtt{/C}$ &  & 3.5 & 3.1 & 17.0 & \hl{18.8} & 3.2 & 5.4  \tabularnewline
$\mathtt{/TNBC}$ &  & 0.7 & \ul{1.2} & 6.5 & \hl{9.3} & \ul{2.9} & \ul{2.3} \tabularnewline
\vspace{-1em}
 &  &  &  &  &  &  &\tabularnewline
\cline{1-1} \cline{3-8} 
\end{tabular}
\par\end{centering}
\vspace{1em}

\caption{\label{tab:supp-multiarch-game}Extended version of table \ref{tab:adv-counter} in the main paper for all four network architectures. Recogniser's payoff table $p_{ij}$, $i\in \Theta^u$, $j\in\Theta^r$, for various convnet architectures. The user's payoff is given by $100-p_{ij}$.}
\end{table}

\newpage

\end{document}